\documentclass[journal]{IEEEtran}

\usepackage{siunitx}
\usepackage{tabularx} 
\usepackage{amsmath}  
\usepackage{graphicx} 
\usepackage{placeins}
\usepackage[draft]{pgf}
\usepackage{cite} 
\usepackage{subfiles}
\usepackage[final]{hyperref} 
\hypersetup{
	colorlinks=true,       
	linkcolor=blue,        
	citecolor=blue,        
	filecolor=blue,     
	urlcolor=blue
}
\usepackage{float}
\usepackage[outdir=./]{epstopdf}
\usepackage{booktabs}
\usepackage{multirow}
\usepackage[english]{babel}
\usepackage[utf8]{inputenc}
\usepackage[capitalise]{cleveref} 
\usepackage[ruled, linesnumbered]{algorithm2e}
\usepackage[noend]{algpseudocode}
\usepackage{hyperref}
\usepackage{fmtcount} 
\usepackage{listings}
\usepackage{caption}
\usepackage{subcaption}
\usepackage{color}
\usepackage{soul}
\usepackage{tikz}
\usepackage{listings}
\usepackage{amsfonts}

\definecolor{darkred}{rgb}{0.55, 0.0, 0.0}
\definecolor{darkspringgreen}{rgb}{0.09, 0.45, 0.27}
\definecolor{steelblue}{rgb}{0.27, 0.51, 0.71}

\newcommand{\blue}[1]{\textcolor{blue}{#1}}

\DeclareMathOperator{\hadamardtimes}{\odot}

\newcommand\revised[1]{#1}

\newcommand{\etal}[1]{#1~\textit{et al.}}
\newcommand{\figref}[1]{Fig.~\ref{#1}}
\newcommand{\equref}[1]{Eq.~(\ref{#1})}
\newcommand{\tabref}[1]{Table~\ref{#1}}

\ifCLASSINFOpdf
\else
\fi
\begin{document}
	%
	\title{Light Field View Synthesis via Aperture Disparity and Warping Confidence Map}
	%
	%
	%
	
	\author{Nan~Meng, 
		Kai~Li, 
		Jianzhuang~Liu,~\IEEEmembership{Senior Member,~IEEE}
		and~Edmund~Y.~Lam,~\IEEEmembership{Fellow,~IEEE}
		\thanks{Part of the work was done during an internship at Huawei. 
		} 
		\thanks{Nan~Meng, and Edmund Y.~Lam are with the Department of Electrical and Electronic Engineering, The University of Hong Kong, Pokfulam, Hong Kong (e-mail: nanmeng@hku.hk, elam@eee.hku.hk)}
		\thanks{Kai~Li is with  Shanghai Jiao Tong University (kai.li@sjtu.edu.cn)} 
		\thanks{Jianzhuang~Liu is with the Noah's Ark Lab, Huawei Technologies Company Limited, Shenzhen 518129, China (liu.jianzhuang@huawei.com)} 
	}
	
	%
	%

	\markboth{Journal of \LaTeX\ Class Files,~Vol.~14, No.~8, August~2015}%
	{Shell \MakeLowercase{\textit{et al.}}: Bare Demo of IEEEtran.cls for IEEE Journals}
	%



	\maketitle
	
	\begin{abstract}
This paper presents a learning-based approach to synthesize the view from an arbitrary camera position given a sparse set of images. A key challenge for this novel view synthesis arises from the reconstruction process, when the views from different input images may not be consistent due to obstruction in the light path. We overcome this by jointly modeling the epipolar property and occlusion in designing a convolutional neural network. We start by defining and computing the aperture disparity map,
which approximates the parallax and measures the pixel-wise shift between two views. While this relates to free-space rendering and can fail near the object boundaries, we further develop a warping confidence map to address pixel occlusion in these challenging regions. The proposed method is evaluated on diverse real-world and synthetic light field scenes, and it shows better performance over several state-of-the-art techniques.



	\end{abstract}
	

	%
	\IEEEpeerreviewmaketitle

	%
	%
	%
	%
	
	\section{Introduction}
	\label{sec:introduction}
	Perceiving the 3-dimensional (3D) nature of an object is a basic human instinct, but it can be very difficult for a computer. One reason is that the optical system cannot easily capture the geometric information of the scene, and therefore it often has difficulty recreating the visual perception~\cite{Lam2015Computational}.
Ordinarily, an image is not informative about the differences among light rays coming from various directions, as they are combined together to form the intensity at a pixel location.
These differences, however, are crucial for us to perceive the world~\cite{Ng2006Digital}. 

With advanced techniques and 
imaging setups,
recording the 3D information of an object is becoming feasible~\cite{Chen20183D}. One of the most promising techniques is light field photography, which uses a plenoptic camera to capture both the directions and radiance of the incident light rays~\cite{Ng2005Light}. The additional directional information allows a wider range of vision applications, such as depth estimation~\cite{Sun2016Data,Schilling2018Trust}, rendering~\cite{Chai2000Plenoptic,Meng2019Computational}, refocusing~\cite{Vigano2019Advanced}, super-resolution~\cite{Meng2019Spatial,Wang2019Spatial} etc. 
However, there is a limit to the density of the pixels that one can capture, necessitating a trade-off between spatial and angular resolution~\cite{Meng2019Highdimensional}.

One way to relieve such a trade-off is to produce the intermediate views from the captured images. 
Ordinarily, approaches for view synthesis require geometric knowledge as priors in the recovery of a dense light field from the sparsely-sampled inputs~\cite{Chaurasia2013Depth,Overbeck2018System}. The new view can be reconstructed from any acquired images by projecting the pixels to proper 3D locations and re-projecting them onto the target picture~\cite{Chai2000Plenoptic}.
However, such methods struggle with complex regions such as occlusions, as well as transparent and semi-transparent surfaces, where the depth information is difficult to calculate or estimate. Without accurate geometry information, 
the generated results often contain
jarring artifacts~\cite{Wu2017LightJSTSP,Kalantari2016Learning}.

An alternative approach is image-based rendering (IBR), which directly reuses the pixels from the available images to produce the new views. For light field, the depth information is not necessary for the rendering process~\cite{Shum2000Review}. The spatio-angular redundancy makes it possible to infer the novel view from neighboring sub-aperture images (SAIs). One advantage of  the IBR techniques is that they can avoid explicitly modeling the realistic geometry of the scene. However, such approaches usually require more samples to counter the undesirable aliasing effects in the outputs~\cite{Chai2000Plenoptic,Zhou2016View}.

These difficulties recently have led to investigations using the learning-based approach, which forgoes the explicit modeling of the problem and instead makes use of deep learning to approximate the ground truth with many training samples~\cite{Flynn2016Deepstereo,Wu2017LightJSTSP}. The powerful representation ability of convolutional neural network (CNN) and its widespread success in vision tasks make this a promising direction
~\cite{Meng2018Large,Sun2016Sparse}. Generally, the CNN-based view synthesis algorithms are able to achieve relatively high-quality reconstruction, but often require the ground truth to acquire the accurate supervision signal~\cite{Wu2018Light,Meng2019Highdimensional}, which restricts the generalization or capacity of the algorithms. For instance, \etal{Meng}~\cite{Meng2020Highorder} and \etal{Yeung}~\cite{Yeung2018Fast} both adopt the learning framework to directly approximate the ground truth. However, the drawback is that they can only generate the views that are recorded in the labels.

Recently, multi-plane image (MPI), which is originally proposed as a scene representation for stereo imagery~\cite{Zhou2018Stereo}, has attracted increasing attention in light field imaging~\cite{Duvall2019Compositing,Flynn2019Deepview}. One example is the pipeline proposed by \etal{Mildenhall}~\cite{Mildenhall2019Local}, which learns the MPI representations of each input image with a 3D CNN. They are subsequently warped and blended to reconstruct the target view. Meanwhile, a similar framework is developed by \etal{Flynn}~\cite{Flynn2019Deepview}, but they adopt learned gradient descent to extract the MPIs. A major benefit of such a layered representation is that it can be reused to reconstruct multiple views at arbitrary camera positions.

The idea of continuous view synthesis has also been explored in video frame interpolation~\cite{Bao2019Depth}. Typically, one estimates the optical flow and uses it to warp input frames to produce the interpolation samples~\cite{Liu2019Deep}. For light field, the warping is usually based on the estimated disparity map. A recent method utilizes fused-pixel and feature-based (a.k.a. FPFR)~\cite{Shi2020Learning} information follows this idea, but the difference is that it warps both the input images and features. In doing so, the model consists of two branches, i.e. one for pixel-based reconstruction and the other for feature-based reconstruction, and the target image is synthesized by combining the intermediate outputs from both.

In this work, we focus on the inherent epipolar property of the light field, and explicitly model the relations among the disparity maps obtained from various SAIs.
\revised{According to the structural property of the light field, we approximate the relationship between disparity value and changes in the angular viewpoint position with a linear relation.
This assumption} allows us to calculate the intermediate disparity map between the input and target views, which can be further exploited to warp the input images.
We name the matrix measuring the pixel shift between different views of light field the \textit{aperture disparity map} (ADM) to emphasize \revised{that there is a relationship} between the value of ADM and the aperture position. \revised{However, in practice, 
such linearity assumption may not always hold for real-world light fields, due to many physical and environmental factors, such as lens distortion of the camera, chaotic environmental light rays, and non-Lambertian object surfaces.}
To \revised{compensate for the defects and} address the pixel occlusion issue, we further estimate the warping confidence maps (WCMs), which equilibrate the radiance information from different input views, to produce the final image. Finally, the coarse results are further refined by a four-dimensional (4D) CNN with alternating filters~\cite{Yeung2018Fast}.

In summary, the contributions of this paper are as follows:
\begin{itemize}
    \item We propose a novel disparity model known as ADM, that is tailored to the light field images to measure the pixel-wise shift distance between a given pair of images. Experimental results show that it works well on both real-world and synthetic scenes.

	\item We introduce the WCM to combine the pixel values from different views for the target image. It can efficiently handle occluded pixels, and therefore reduce the artifacts near the object boundaries.
\end{itemize}

	\section{Related work}
	\label{sec:related_work}
	View synthesis and IBR are closely related. Generally, the explicit geometry models are not necessary for IBR when generating new images in a light field, while view synthesis usually requires both geometry information and a few images to provide the virtual views. 
In this section, we review the literature of IBR algorithms, view synthesis, and recent learning-based approaches in the context of light field imaging.

Early IBR algorithms for light field rely on the characterization of the plenoptic function and treat the creation of new views as resampling~\cite{Levoy2006Lightfields,Kubota2006Reconstructing}. This approach ignores  occlusion, and thus is only feasible for free space rendering or for producing the views reasonably close to the original ones. It gradually becomes clear that interpolation of plausible views in high quality requires either intensive sampling or  knowledge about the scene~\cite{Chai2000Plenoptic,Lin2004Geometric}. Therefore, a different set of approaches to light field rendering attempt to infer at least some geometry information, which include methods that rely on image registration~\cite{Siu2005Image}, prior knowledge~\cite{Fitzgibbon2005Image}, or image correspondence and warping~\cite{Seitz1996View}, to name a few.

View synthesis methods, on the other hand, focus on making use of explicit geometric knowledge that assists in the recovery of a dense light field. Some enforce explicit priors on the light field itself, such as  sparsity  in the Fourier domain~\cite{Shi2014Light} or shearlet transform domain~\cite{Vagharshakyan2018Light}, a patch-based Gaussian mixture model~\cite{Mitra2012Light}, or Lambertian surfaces with modest depth discontinuities~\cite{Levin2010Linear}. However, these methods require either a specific sampling pattern or a large number of views, which limit their practical uses. Other techniques involve partial reconstruction of the scene geometry, such as a global 3D reconstruction~\cite{Hedman2017Casual} or a soft model of the geometrical relationships~\cite{Penner2017Soft}.
Some methods infer the geometry by estimating the disparity for a single view~\cite{Wanner2014Variational} or for each input view~\cite{Overbeck2018System}. Given that an accurate depth estimation is hard to obtain, such  approaches often struggle with complex scenes.

Another group of synthesis algorithms do not require explicit geometric models but rely on the feature correspondence between images. The classical approaches of this kind interpolate the intermediate views by exploiting the optical flow~\cite{Chen1993View,Niklaus2017Video}. However, given that non-Lambertian surfaces and occlusions are still challenging to the flow estimation methods~\cite{Butler2012Naturalistic,Dosovitskiy2015Flownet}, the interpolated views tend to have artifacts near the object boundaries. \etal{Wang}~\cite{Wang2017Light} make use of the images taken from another standard camera as references to generate plausible frames for the light field video, which also increases the complexity of the imaging system.

More recently, learning-based methods come into the spotlight due to their effectiveness on vision tasks~\cite{Meng2018Large,Meng2020Lightgan}. According to the way learning is involved, such methods can generally be divided into two groups. The first one attempts to establish a direct mapping using learning frameworks between the sparsely-sampled inputs and their dense correspondence. The difference among various techniques lies in the reconstruction level. For instance, Gul and Gunturk~\cite{Gul2018Spatial} handle the pixel-wise reconstruction. LFCNN~\cite{Yoon2017Light} and \etal{Wang}~\cite{Wang2018Lfnet} are aperture-wise methods. \etal{Wu}~\cite{Wu2018Light} explore an approach to recover the epipolar plane image (EPI), while \etal{Meng}~\cite{Meng2019Highdimensional} and \etal{Yeung}~\cite{Yeung2018Fast} directly restore the entire light field. The second group embeds learning in the traditional rendering pipeline. \etal{Kalantari}~\cite{Kalantari2016Learning} make one of the early attempts to adopt two CNNs for disparity estimation and color prediction. Meanwhile, \etal{Srinivasan}~\cite{Srinivasan2017Learning} propose a two-stage learning process to estimate scene geometry and eliminate occlusions.

Generally, the learning-based methods produce more plausible visual results, but they also require a large amount of training data paired with labels. To overcome such a problem, \etal{Chen}~\cite{Chen2020Self} come up with a self-supervised approach by fine-tuning a video interpolation framework based on cycle consistency. \etal{Gao}~\cite{Gao2020Self} employ a CNN to restore EPI coefficients in the shearlet domain. In addition to the data volume, for many prevailing learning methods, the rigid learning strategy heavily relies on the training data, consequently restricting generalization of the model. The end-to-end training pattern makes many models hard to reconstruct an image from a viewpoint that has not been recorded in the ground truth~\cite{Yeung2018Fast,Meng2020Highorder}.

	\section{Method}
	\label{sec:method}
We adopt the two-plane parametrization~\cite{Levoy1996Light} to represent the 4D light field. Each light ray is represented by the intersections with two parallel planes transmitting from the spatial coordinate $\boldsymbol{x} = (x,y)$ to the angular coordinate $\boldsymbol{u}=(u, v)$, and thus denoted by $L := L(\boldsymbol{x},\boldsymbol{u})$. We assume that all the coordinate variables in the plenoptic function $L(\boldsymbol{x},\boldsymbol{u})$ are continuous. Therefore, the goal of IBR is to reconstruct such a function based on a set of discrete samples $L(\boldsymbol{x}_i,\boldsymbol{u}_j)$ ($i, j \in \mathbb{N}$),


\begin{equation}\label{equ:IBR}
L(\boldsymbol{x}_i,\boldsymbol{u}_j) \xrightarrow{g} L(\boldsymbol{x},\boldsymbol{u}),
\end{equation}
where $g(\cdot)$ denotes the reconstruction algorithm.

\subsection{Spatio-Angular Relationship}
\begin{figure}[t]
	\centering
	\includegraphics[width=.72\columnwidth]{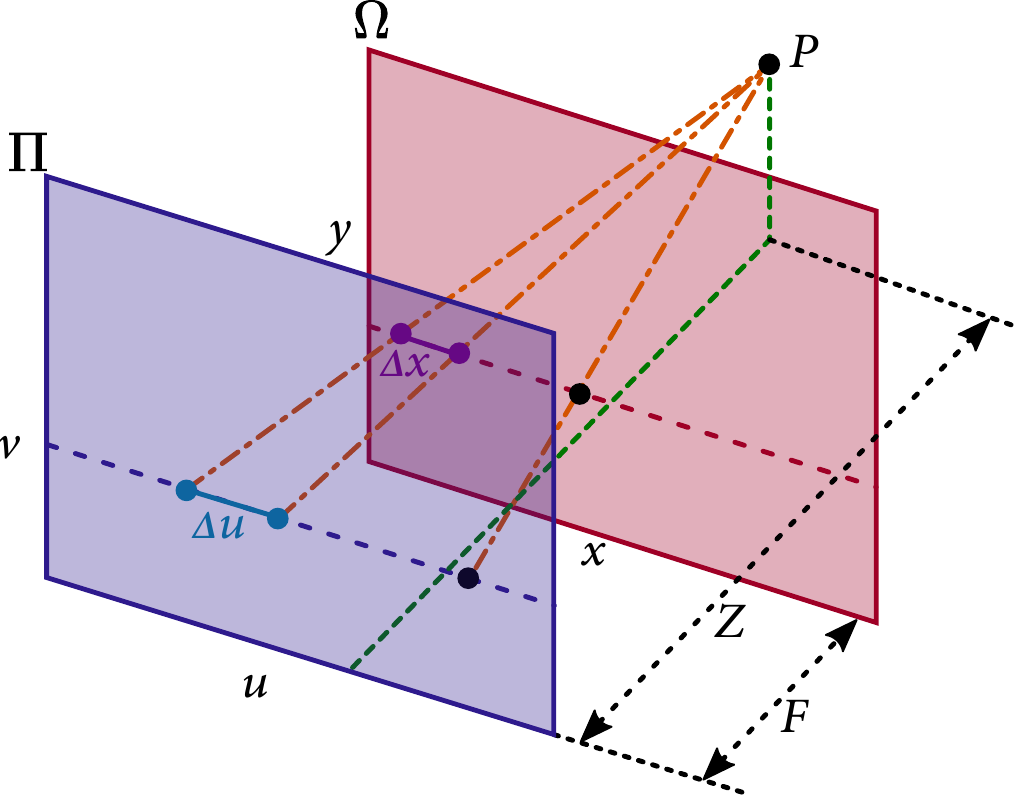}
	\caption{Illustration of pixel shift from different viewpoints in the two-plane parametrization of light field imaging.}
	\label{fig:spatio_angular_relation}
\end{figure}

\begin{figure*}[!ht]
	\centering
	\begin{subfigure}[t]{0.32\textwidth}
		\centering
		\includegraphics[width=1.\textwidth]{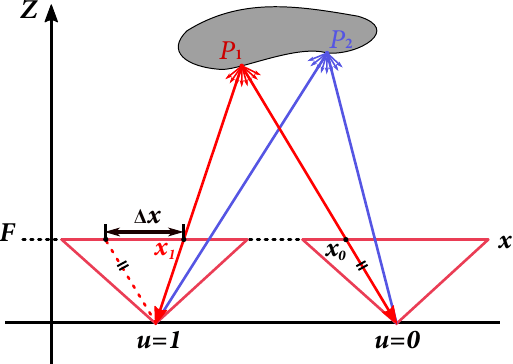}
		\caption{Illustration of light ray transmission in free space.
		}
		\label{subfig:free_space_transmission}
	\end{subfigure}
	\hfill
	\begin{subfigure}[t]{0.32\textwidth}
		\centering
		\includegraphics[width=1.\textwidth]{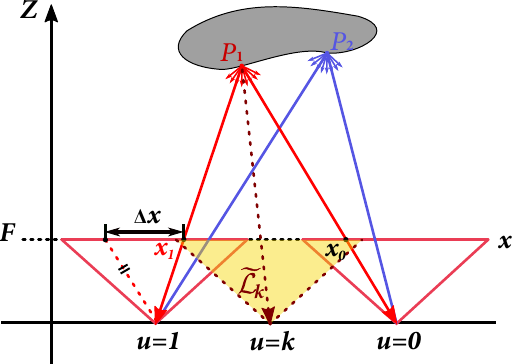}
		\caption{Illustration of intermediate radiance inference model in free space.}
		\label{subfig:free_space_inference}
	\end{subfigure}
	\hfill
	\begin{subfigure}[t]{0.32\textwidth}
		\centering
		\includegraphics[width=1.\textwidth]{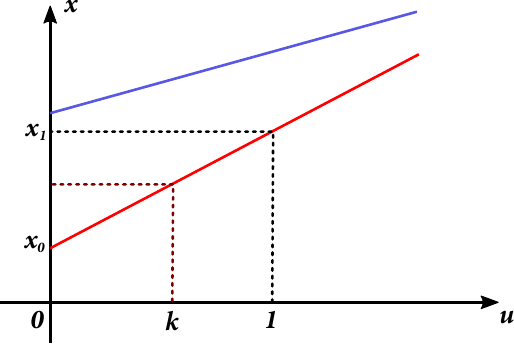}
		\caption{EPI pattern corresponding to (b). Each line has a uniform color due to the photo-consistency assumption.}
		\label{subfig:free_space_epi}
	\end{subfigure}
	\caption{Intermediate radiance inference model in \revised{space free of occluders in light field}.
	}
	\label{fig:radiance_inference}
\end{figure*}
For any given point $P$ in the free space, the two-plane parametrization is depicted in Fig.~\ref{fig:spatio_angular_relation}, where $F$ is the orthogonal distance between the two parallal planes, and $Z$ is the depth of the focused point. The pixel shifts in different views can be inferred using similar triangles. If we vary $\boldsymbol{x}$ with a distance $\Delta \boldsymbol{x}$, the angular coordinate has to change according to
\begin{equation}\label{equ:shift}
\Delta \boldsymbol{x} = \frac{Z-F}{Z} \Delta \boldsymbol{u} = \frac{\alpha F-F}{\alpha F} \Delta \boldsymbol{u}
=\left(1 - \frac{1}{\alpha}\right) \Delta \boldsymbol{u},
\end{equation}
where $\alpha=\frac{Z}{F}$ denotes the disparity ratio. $\Delta \boldsymbol{u}$ is the distance between two viewpoints located at the camera plane $\Pi$.
In the following, we pay more attention to the appearance of the view, and for the sake of illustration, we use the symbol $\mathcal{L}_{\boldsymbol{u}}(\cdot)$ to denote the plenoptic function obtained from the viewpoint at location $\boldsymbol{u}$ as
\begin{equation}\label{equ:simple_plenoptic_function}
\mathcal{L}_{\boldsymbol{u}}(\boldsymbol{x}) := L(\boldsymbol{x}, \boldsymbol{u}).
\end{equation}
An equivalent expression of \equref{equ:simple_plenoptic_function} is
\begin{equation}\label{equ:identical}
\mathcal{L}_{\boldsymbol{u}}(\boldsymbol{x}) = \mathcal{L}_{\boldsymbol{u+\Delta u}}(\boldsymbol{x}+\boldsymbol{\Delta x}),
\end{equation}
if we assume the pixels in an image shift by $\boldsymbol{\Delta x}$ when the viewpoint changes by $\boldsymbol{\Delta u}$.

Substituting \equref{equ:shift} into \equref{equ:identical}, we obtain
\begin{equation}
\begin{aligned}\label{equ:SAI_relations}
\mathcal{L}_{\boldsymbol{u}}(\boldsymbol{x}) &= \mathcal{L}_{\boldsymbol{u+\Delta u}}(\boldsymbol{x}+\boldsymbol{\Delta x}) \\
&= \mathcal{L}_{\boldsymbol{u+\Delta u}}\left( \boldsymbol{x}+ \left( 1-\frac{1}{\alpha} \right)\boldsymbol{\Delta u}\right).
\end{aligned}
\end{equation}

\subsection{Photo-Consistency}
\label{sec3:photo_consistency}
The photo-consistency assumption is commonly adopted in the multi-view vision tasks. It assumes that all light rays coming from the same focus point in the scene should result in the same photometric values. Since the rays from different directions are recorded separately in a light field, this assumption means that the value of the recorded pixels in different SAIs  corresponding to the same point of the scene should be identical. Nevertheless, it does not always hold, and usually fails when there are occlusions along the path of light ray transmission.

\subsection{Free-Space Intermediate Radiance Inference}
\label{sec3:intermediate_radiance_inference_freespace}
We first establish a model to describe the radiance of a light ray impinging on an arbitrary position in the camera plane. Assume that all the light rays are transmitted in free space, \revised{i.e., the space free of occluders}, as illustrated in \figref{subfig:free_space_transmission}. For simplicity, in the following we will retain only one angular coordinate $u$, and derive $L(\boldsymbol{x}, u)$ for a collection of light rays traveling from the position $u$ to the position $\boldsymbol{x}$ in the respective planes. Such collection of light rays form the view image from the viewpoint $u$. We consider the continuous change of viewpoint $u$ and set its range to be $[0, 1]$. The two boundary viewpoint positions are $u=0$ and $u=1$.

Given two collections of light rays $\mathcal{L}_0(\boldsymbol{x})$ and $\mathcal{L}_1(\boldsymbol{x})$ from two different viewpoints, and a factor $k \in (0, 1)$, our goal is to infer the intermediate rays $\widetilde{\mathcal{L}}_k(\boldsymbol{x})$, as is demonstrated in \figref{subfig:free_space_inference}.
\revised{According to \equref{equ:shift}, 
one can
approximate the relationship between coordinates $u$ and $x$ using a linear mapping.}
The EPI is a map obtained by gathering the light field samples with a fixed spatial ($x$ or $y$) and a fixed angular ($u$ or $v$) coordinate. It can reflect the relationship between a pair of spatial and angular coordinates ($x$ and $u$, or $y$ and $v$). Given a certain point on the Lambertian surface with depth $Z$, when changing the viewpoint $u$, the spatial position $x$ will also change according to \equref{equ:shift}, forming a line on the EPI. The slope of the line is related to the depth of the point.

\figref{subfig:free_space_epi} shows the EPI pattern of two points at different depths. The red line corresponds to the point $P_1$ while the blue line corresponds to $P_2$.
The photo-consistency assumption ensures that each line should have a uniform color, i.e. projections of the same point in different views should have the same intensity value.
This allows us to approximate the radiance $\mathcal{L}_k(\boldsymbol{x})$ by shifting the pixels corresponding to $\mathcal{L}_0(\boldsymbol{x})$ and $\mathcal{L}_1(\boldsymbol{x})$ properly. Also, in terms of \equref{equ:SAI_relations}, we have 
\begin{equation}\label{equ:Lk_estimation}
\begin{alignedat}{3}
\widetilde{\mathcal{L}}_k(\boldsymbol{x}) &\cong (1-k) \cdot \mathcal{L}_k(\boldsymbol{x}) + k \cdot \mathcal{L}_k(\boldsymbol{x}) \\
&= (1-k) \cdot \mathcal{L}_0\left( \boldsymbol{x}-\left( 1-\frac{1}{\alpha} \right)k \right)+ \\
&\qquad k  \cdot \mathcal{L}_1\left( \boldsymbol{x}+\left( 1-\frac{1}{\alpha} \right)(1-k) \right),
\end{alignedat}
\end{equation}
where $\widetilde{\mathcal{L}}_k(\boldsymbol{x})$ denotes the estimate of $\mathcal{L}_k(\boldsymbol{x})$. \equref{equ:Lk_estimation} provides a more general expression. Practically, $\mathcal{L}_0(\boldsymbol{\boldsymbol{x}})$ and $\mathcal{L}_1(\boldsymbol{x})$ are usually close but not the same, due to the noise and illumination. Therefore, the coefficient $k$ can also be regarded as a weighting factor to balance the information from $\mathcal{L}_0(\boldsymbol{x})$ and $\mathcal{L}_1(\boldsymbol{x})$ for the estimate. However, since the depth information of the radiance is unknown, the ratio $\alpha$ cannot be computed.

To mitigate such a problem, we develop a way to estimate the disparity ($\boldsymbol{\Delta x}$) directly. An ADM is learned by a dense CNN, which provides the pixel-wise shift information between a pair of view images. 
Here, we use it to obtain the radiance inference. A more detailed demonstration will be presented in Section~\ref{sec3:aperture_optical_flow_estimation}.

Mathematically, $A_{k \leftarrow 0}(\boldsymbol{x})$ and $A_{k \leftarrow 1}(\boldsymbol{x})$ denote the ADMs from $\mathcal{L}_0(\boldsymbol{x})$ to $\mathcal{L}_k(\boldsymbol{x})$ and $\mathcal{L}_1(\boldsymbol{x})$ to $\mathcal{L}_k(\boldsymbol{x})$, respectively.
Following \equref{equ:Lk_estimation}, we have
\begin{equation}\label{equ:inference}
\widetilde{\mathcal{L}}_k(\boldsymbol{x}) = (1-k) P \Big( \mathcal{L}_0(\boldsymbol{x}), A_{k \leftarrow 0}(\boldsymbol{x}) \Big) + k P \Big( \mathcal{L}_1(\boldsymbol{x}), A_{k \leftarrow 1}(\boldsymbol{x}) \Big),
\end{equation}
where $P(\cdot, \cdot)$ is the pixel-wise warping operation that translates each pixel of the input image to its correspondence in the target image in terms of the disparity map between the two images. Given the input image $\mathcal{L}_i(\boldsymbol{x})$ ($i \in [0, 1]$) and the ADM $A_{o \leftarrow i}(\boldsymbol{x})$ ($o \in [0, 1]$), the output image of the function $P(\cdot, \cdot)$ is
\begin{equation}
P \Big(\mathcal{L}_i(\boldsymbol{x}), A_{o \leftarrow i}(\boldsymbol{x}) \Big) = \mathcal{L}_o(\boldsymbol{x}) = \mathcal{L}_i \Big(\boldsymbol{x} + A_{o \leftarrow i}(\boldsymbol{x}) \Big).
\end{equation}
\equref{equ:inference} implies that the closer the viewpoint position $k$ is to $0$, the more contribution $\mathcal{L}_0(\boldsymbol{x})$ will make to $\widetilde{\mathcal{L}}_k(\boldsymbol{x})$.

\subsection{Inference with Occlusions}
\label{sec3:intermediate_radiance_inference_with_occlusions}

\begin{figure}[t]
	\centering
	\includegraphics[width=0.98\columnwidth]{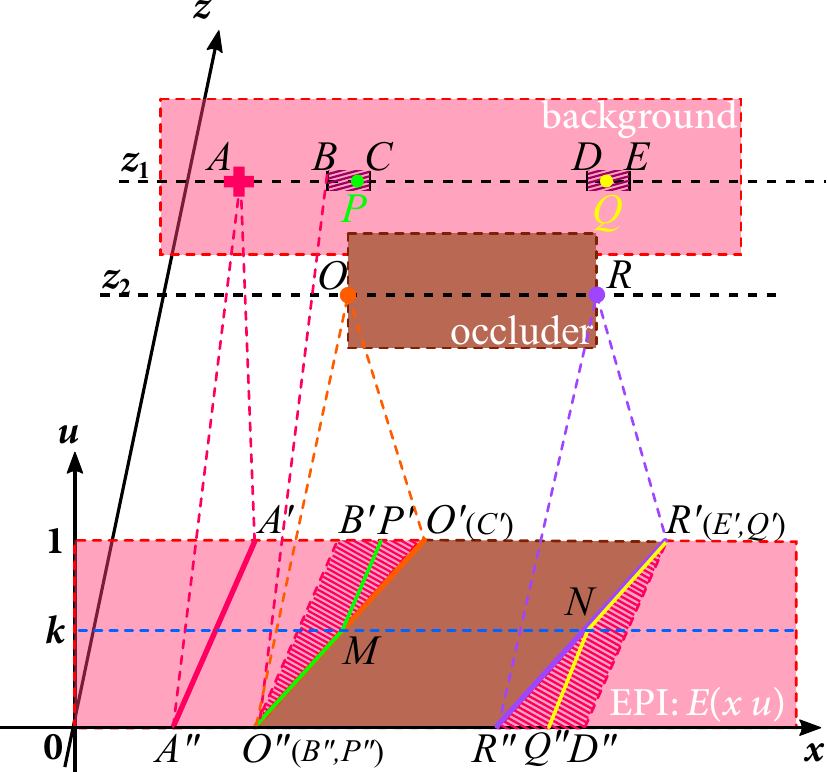}
	\caption{EPI pattern with occlusions between the observer and the object.}
	\label{fig:occlusions_analysis}
\end{figure}
We next discuss the situation where there exist occlusions along the path of the light ray, violating the photo-consistency assumption.
Given the direction of light rays recorded by a plenoptic camera, an important property of the light field  is that the \emph{occluded pixel in the synthesized image always appears in either the leftmost SAI or the rightmost SAI (i.e., the boundary images)}.
This is illustrated in Fig.~\ref{fig:occlusions_analysis}, which presents the EPI pattern of the dashed line region of two objects placed at different distances from the camera. The near object (named ``occluder'') with depth $z_2$ partially occludes the further object (named ``background'') with distance $z_1$. On the background, the region $CD$ is totally occluded in all views. The regions $BC$ and $DE$ are partially occluded and the other places are totally exposed. In the EPI, each slope line corresponds to a point. For example, the line $A'A''$ with a small slope on EPI corresponds to the point $A$ of the background, while the line $O'O''$ is projected from point $O$ on the occluder with a larger slope on the EPI pattern.

With such a configuration, we now illustrate how to modify the free-space inference model to fit scenarios with occlusions. As the occlusion appears near the boundary of the object, we shade the corresponding EPI in Fig.~\ref{fig:occlusions_analysis} to highlight these regions with occlusions, i.e. $B'O'O''$ and $R'D''R''$.
We first focus on a certain pixel located at point $P$ on the background that is also at the boundary of the occluder when $u=k$. Its EPI pattern is marked with a green line $P'MP''$. This point is occluded when $u \in (0, k)$, but is exposed when $u \in (k, 1)$. On the other hand, another point $Q$ is projected to the line $Q'NQ''$, which is occluded when $u \in (k, 1)$ but is exposed when $u \in (0, k)$. 

Nevertheless, both $P$ and $Q$ can be inferred from one of the boundary images, i.e. $\mathcal{L}_0(\boldsymbol{x})$ for $Q$ and $\mathcal{L}_1(\boldsymbol{x})$ for $P$. As a consequence, when there are occlusions along the light ray, the missing pixel information can be inferred from one of the boundary images. Based on this observation, we design another network to estimate the warping confidence maps, known as WCMs, to assist in the inference of the pixel value. The estimated WCMs, $O_{k \leftarrow 0}(\boldsymbol{x})$ and $O_{k \leftarrow 1}(\boldsymbol{x})$, denote the confidence level that a pixel value of $\mathcal{L}_k(\boldsymbol{x})$ can be inferred from $\mathcal{L}_0(\boldsymbol{x})$ and $\mathcal{L}_1(\boldsymbol{x})$, respectively.

As a result, Eq.~\ref{equ:inference} can be modified as
\begin{equation}
\begin{aligned}\label{equ:inference_with_occlusion}
\widetilde{\mathcal{L}}_k(\boldsymbol{x}) = \Phi^{-1} \hadamardtimes \Big[ &(1-k)O_{k \leftarrow 0}(\boldsymbol{x}) \hadamardtimes P \Big( \mathcal{L}_0(\boldsymbol{x}), A_{k \leftarrow 0}(\boldsymbol{x}) \Big) + \\
&k O_{k \leftarrow 1}(\boldsymbol{x}) \hadamardtimes P \Big( \mathcal{L}_1(\boldsymbol{x}), A_{k \leftarrow 1}(\boldsymbol{x}) \Big) \Big],
\end{aligned}
\end{equation}
where $\Phi = k O_{k \leftarrow 0}(\boldsymbol{x}) + (1-k)O_{k \leftarrow 1}(\boldsymbol{x})$ denotes a normalization factor. The symbol $\hadamardtimes$ denotes the Hadamard product between two matrices. The values of the confidence map should fall within $[0, 1]$. Take, as an example, the point $P$ (with coordinate $\boldsymbol{p}$), which is occluded when $u=0$ and exposed when $u=1$. We should have $O_{k \leftarrow 0}(\boldsymbol{p}) = 0$, which means that $\mathcal{L}_0(\boldsymbol{p})$ has no contribution to  $\widetilde{\mathcal{L}}_k(\boldsymbol{p})$, and $O_{k \leftarrow 1}(\boldsymbol{p}) = 1$, such that the value is fully contributed by $\mathcal{L}_1(\boldsymbol{p})$. Similarly, for point $Q$ (with coordinate $\boldsymbol{q}$), the value of two confidence maps should satisfy $O_{k \leftarrow 0}(\boldsymbol{q}) = 1$ and $O_{k \leftarrow 1}(\boldsymbol{q}) = 0$. For the points not in the partially occluded region, the value is contributed from both boundary light rays. Consequently, these two maps should satisfy the constraint
\begin{equation}\label{equ:confidence_map_relation}
O_{k \leftarrow 0}(\boldsymbol{x}) + O_{k \leftarrow 1}(\boldsymbol{x}) = 1.
\end{equation}

\subsection{Aperture Disparity Map Estimation}
\label{sec3:aperture_optical_flow_estimation}
\begin{figure}[t]
	\centering
	\includegraphics[width=0.8\columnwidth]{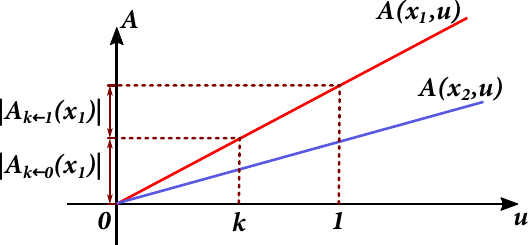}
	\caption{Illustration of a linear relation between $A(\cdot, \cdot)$ and the angular coordinate $u$. The values $x_1$ and $x_2$ denote the pixels projected from points $P_1$ and $P_2$ (\figref{fig:radiance_inference}), respectively.
	The EPI patterns corresponding to the two points are highlighted in different colors.}
	\label{fig:relation_aperture_flow}
\end{figure}
After we obtain the occlusion-aware inference expression in \equref{equ:inference_with_occlusion}, the next crucial problem is how to estimate the intermediate ADM $A_{k \leftarrow 1}(\boldsymbol{x})$ and $A_{k \leftarrow 0}(\boldsymbol{x})$ for target view synthesis. As discussed earlier in Section~\ref{sec3:intermediate_radiance_inference_freespace}, each value of the ADM represents the spatial shift of the corresponding pixel (radiance). \revised{It approximates the relationship between shift distance and viewpoint changes using a linear mapping.}
To demonstrate this, 
we define an auxiliary variable $A(\boldsymbol{x}, u)$, which denotes the distance $\boldsymbol{x}$ has shifted from the original viewpoint ($0$) to viewpoint $u$. Consequently, ADM can also be expressed using the auxiliary variable, i.e. $A(\boldsymbol{x}, u) = A_{u \leftarrow 0}(\boldsymbol{x})$.

According to \equref{equ:shift}, one can easily obtain the proportional relation between $A(\cdot, \cdot)$ and coordinate $u$, as demonstrated in \figref{fig:relation_aperture_flow}. Therefore, for any $\boldsymbol{x}$ and $k$, the disparity value satisfies
\begin{align}
A(\boldsymbol{x}, 0) &= 0 \label{equ:property1} \\
A_{k \leftarrow 0}(\boldsymbol{x}) &= A(\boldsymbol{x}, k) - A(\boldsymbol{x}, 0) = -A_{0 \leftarrow k}(\boldsymbol{x}) \label{equ:property2}\\
A_{k \leftarrow 1}(\boldsymbol{x}) &= A(\boldsymbol{x}, k) - A(x, 1) = -A_{1 \leftarrow k}(\boldsymbol{x}) \label{equ:property3}\\
A_{k \leftarrow 0}(\boldsymbol{x}) &= k \cdot A_{1 \leftarrow 0}(\boldsymbol{x}) = -k \cdot A_{0 \leftarrow 1}(\boldsymbol{x}) \label{equ:property4}\\
A_{k \leftarrow 1}(\boldsymbol{x}) &= (k-1) \cdot A_{1 \leftarrow 0}(\boldsymbol{x}) = (1-k) \cdot A_{0 \leftarrow 1}(\boldsymbol{x}) \label{equ:property5} \\
A_{1 \leftarrow 0}(\boldsymbol{x}) &= A(\boldsymbol{x}, 1) - A(\boldsymbol{x}, 0) = A_{1 \leftarrow k}(\boldsymbol{x}) + A_{k \leftarrow 0}(\boldsymbol{x}) . \label{equ:property6}
\end{align}
\cref{equ:property1,equ:property2,equ:property3,equ:property4,equ:property5} are derived directly from the epipolar property of light field described by \equref{equ:shift} and \equref{equ:SAI_relations}. In addition, \equref{equ:property6} can be deduced by combining \equref{equ:property2} and \equref{equ:property3}.
Therefore, ADMs $A_{k \leftarrow 1}(\boldsymbol{x})$ and $A_{k \leftarrow 0}(\boldsymbol{x})$ can be calculated in terms of $A_{1 \leftarrow 0}(\boldsymbol{x})$ and $A_{0 \leftarrow 1}(\boldsymbol{x})$ as
\begin{align}
A_{k \leftarrow 0}(\boldsymbol{x}) &= \frac{k+1}{2}A_{1 \leftarrow 0}(\boldsymbol{x}) + \frac{1-k}{2}A_{0 \leftarrow 1}(\boldsymbol{x})\label{equ:Ak0} \\
A_{k \leftarrow 1}(\boldsymbol{x}) &= \frac{k}{2}A_{1 \leftarrow 0}(\boldsymbol{x}) + \left(1-\frac{k}{2}\right)A_{0 \leftarrow 1}(\boldsymbol{x}), \label{equ:Ak1}
\end{align}
if we assume the summation of the weights in Eq.~\ref{equ:Ak0} and Eq.~\ref{equ:Ak1} to be 1. The intermediate radiance inference model derived in Sections~\ref{sec3:intermediate_radiance_inference_freespace} and~\ref{sec3:intermediate_radiance_inference_with_occlusions} can be easily extended to the other angular ($v$) coordinate.

\begin{figure*}[t]
	\centering
	\includegraphics[width=1.\textwidth]{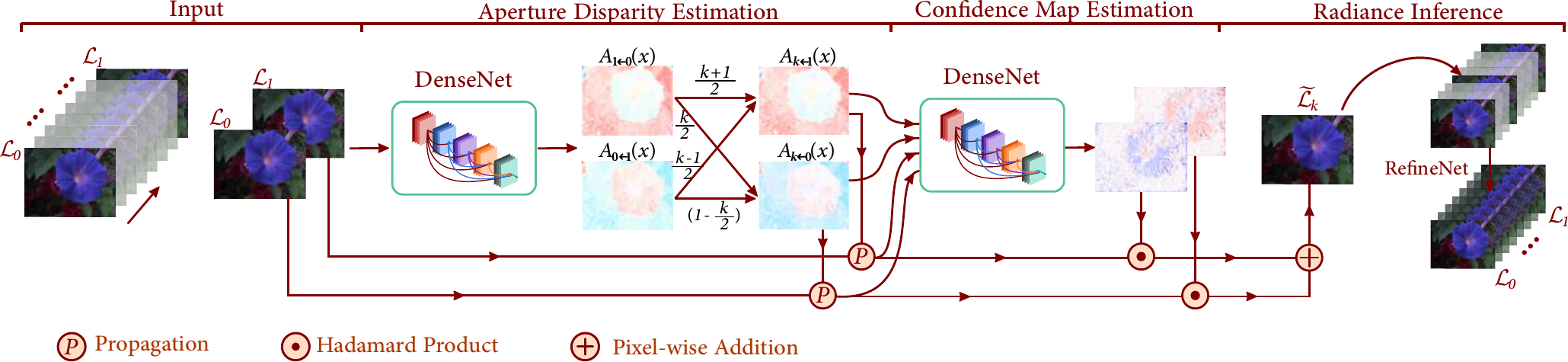}
	\caption{Overview of our proposed framework. Symbol \textcircled{\scriptsize $P$} is the pixel-wise warping operation, $\hadamardtimes$ is the Hadamard product, and $\oplus$ refers to the pixel-wise addition.}
	\label{fig:framework}
\end{figure*}

The expressions in Eq.~\ref{equ:Ak0} and Eq.~\ref{equ:Ak1} allow us to calculate two intermediate ADMs based on the boundary ADMs, $A_{1 \leftarrow 0}(\boldsymbol{x})$ and $A_{0 \leftarrow 1}(\boldsymbol{x})$. Because the boundary images are given, the two boundary ADMs can be generated using deep learning methods. In our framework, ADMs and WCMs are estimated sequentially using a dense network.
As shown in~\cite{Zhang2018Residual}, the dense residual network (RDN) has an effective skip connection pattern, which encourages feature reuse and makes the model more compact and less prone to over-fitting. In addition, each individual layer can directly receive the supervision from the loss function through the shortcut path, which provides implicit deep supervision. Considering such desirable properties of RDN, we adopt it for ADM and WCM estimation.
We use two dense residual blocks, and each includes 4 convolutional layers and 3 ReLU layers. At the end of each block, the learned dense features are concatenated and fed into a convolutional layer with a $1\times1$ kernel to learn more effective features adaptively. In addition, each block allows direct connections from the previous blocks to extract the hierarchical features for disparity and confidence map estimation.

The overview of the proposed framework is shown in \figref{fig:framework}, which mainly consists of three stages (excluding the input). In aperture disparity estimation, our model adopts a dense network to estimate the ADMs between two boundary images. They are further used to approximate the intermediate ADMs, $A_{k \leftarrow 1}(\cdot)$ and $A_{k \leftarrow 0}(\cdot)$. Then, in confidence map estimation, both the intermediate disparity maps and images are fed into a subsequent dense network to obtain the confidence maps that indicate the pixel-wise contribution from the two input SAIs. In radiance inference, the target image at $k$ is computed according to \equref{equ:inference_with_occlusion}. It is further refined by exploiting the parallax information of all the SAIs with the RefineNet. Recently, some studies have demonstrated the effectiveness of 4D convolution filter for spatial details reconstruction. \etal{Meng}~\cite{Meng2019Highdimensional} adopted multiple 4D convolutional layers to recover the high-frequency spatial details. While the results are satisfactory, the framework requires a large amount of computation.
To deal with such a problem, we make use of the alternating convolution filter~\cite{Yeung2018Fast} in our RefineNet. It is a type of 4D convolution filter, which can fully exploit the parallax information of all the views and reduce a significant amount of computation. This fully-convolutional module can then efficiently handle the input light field with changeable angular size. RefineNet contains two 4D filtering steps, and each one is approximated with two alternating filters with kernel size $3\times3\times1\times1$ and $1\times1\times3\times3$, respectively. The learned features are concatenated and then fed into a 4D filter with kernel size $1\times1\times1\times1$ to obtain the refined results.

\subsection{Loss Function}
\label{sec3:loss_function}
All modules and calculations in our framework are differentiable, which enables us to train different parts of our model synchronously. Given the input images $\mathcal{L}_0(\boldsymbol{x})$ and $\mathcal{L}_1(\boldsymbol{x})$, and a set of intermediate images $\{\mathcal{L}_{k_i}(\boldsymbol{x})\}_{i=1}^N$, our loss function consists of four parts. First  is a reconstruction loss that directly provides a supervision signal for the synthesized results by calculating the absolute residual error $\ell_r$ between the intermediate images and the corresponding labels, i.e.,
\begin{equation}\label{equ:absolute_residual_loss}
\ell_r = \frac{1}{N} \sum_{i=1}^N \big \| \mathcal{L}_{k_i}(\boldsymbol{x}) - \widetilde{\mathcal{L}}_{k_i}(\boldsymbol{x}) \big \|_1.
\end{equation}

To reconstruct the high-frequency spatial details~\cite{Ledig2016Photo,Meng2020Lightgan}, we also add the content perceptual loss component $\ell_{c}$ given by
\begin{equation}\label{equ:perceptual_loss}
\ell_{c} = \frac{1}{N} \sum_{i=1}^N \Big \| \phi \Big(\mathcal{L}_{k_i}(\boldsymbol{x}) \Big) - \phi \left(\widetilde{\mathcal{L}}_{k_i}(\boldsymbol{x}) \right)\Big \|^2,
\end{equation}
where $\phi(\cdot)$ maps the input images into high-level feature vectors extracted from the ImageNet pretrained VGG16 model (\texttt{conv4\_3} layer)~\cite{Simonyan2014Very}.

The third part is a warping loss $\ell_{\omega}$, which models the quality of the four estimated AFMs as shown in \figref{fig:framework}. It includes four terms, and each measures the absolute difference between the corresponding propagated views and the ground truth, i.e.,
\begin{equation}\label{equ:propagation_loss}
\begin{aligned}
\ell_{\omega} = 
& \Big \| \mathcal{L}_0(\boldsymbol{x}) - P \big(\mathcal{L}_1(\boldsymbol{x}), A_{0 \leftarrow 1}(\boldsymbol{x}) \big) \Big \|_1 + \\ 
& \Big \| \mathcal{L}_1(\boldsymbol{x}) - P \big(\mathcal{L}_0(\boldsymbol{x}), A_{1 \leftarrow 0}(\boldsymbol{x}) \big) \Big \|_1 + \\
& \frac{1}{N}\sum_{i=1}^N \Big \| \mathcal{L}_{k_i}(\boldsymbol{x}) - P \big(\mathcal{L}_1(\boldsymbol{x}), A_{k_i \leftarrow 1}(\boldsymbol{x}) \big) \Big \|_1 + \\
& \frac{1}{N}\sum_{i=1}^N \Big \| \mathcal{L}_{k_i}(\boldsymbol{x}) - P \big(\mathcal{L}_0(\boldsymbol{x}), A_{k_i \leftarrow 0}(\boldsymbol{x}) \big) \Big \|_1 .
\end{aligned}
\end{equation}
Finally, we smooth the estimated aperture disparity with~\cite{Liu2017Video}
\begin{equation}\label{equ:smooth_loss}
\begin{aligned}
\ell_s = &\left \| \nabla_x A_{0 \leftarrow 1}(\boldsymbol{x}) \right \|_1 + \left \| \nabla_y A_{0 \leftarrow 1}(\boldsymbol{x}) \right \|_1 + \\
&\left \| \nabla_x A_{1 \leftarrow 0}(\boldsymbol{x}) \right \|_1 + \left \| \nabla_y A_{1 \leftarrow 0}(\boldsymbol{x}) \right \|_1 ,
\end{aligned}
\end{equation}
where the notation $\nabla$ denotes the differential operation. 

Our overall objective is a summation of these loss functions, i.e.,
\begin{equation}\label{equ:training_loss}
\ell = \lambda_1 \ell_r + \lambda_2 \ell_{c} + \lambda_3 \ell_{\omega} + \lambda_4 \ell_{s},
\end{equation}
where $\lambda_1$, $\lambda_2$, $\lambda_3$ and $\lambda_4$ are the respective weights. Empirically, they are set to $200$, $1000$, $100$ and $1$, respectively.

	\section{Experiments}
	\label{sec:experiments}
	\subsection{Datasets and Implementation Details}
The effectiveness of data-driven methods often depends significantly on the quality of the training data. Compared with many view synthesis approaches that directly learn to produce the target image, our model attempts to approximate the image relations or correspondences, which increase the model robustness to different types of light field images. To demonstrate this, we train our networks using 100 real-world light field images captured with a Lytro Illum camera provided by the Stanford Lytro Archive~\cite{StanfordLytro}. Due to hardware limitation of plenoptic cameras, many corner angular samples are outside the field of view. Therefore, for each scene, we select the center $9 \times 9$ views in the experiments. To validate the effectiveness of our proposed framework, we conduct experiments on both synthetic and real scenes. The former are selected from HCI datasets~\cite{Honauer2016Dataset,Wanner2013Datasets} generated using the Blender software, while the latter are from multiple public datasets captured with the Lytro Illum cameras, including Stanford Lytro Archive~\cite{StanfordLytro} (not included in the training set), \etal{Kalantari}'s~\cite{Kalantari2016Learning} and Flower~\cite{Srinivasan2017Learning}.

Our project is implemented using PyTorch, and the optimization model is trained on a Ubuntu 16.04.4 computer with an Intel Xeon(R)@2.20HGz CPU and a Titan X GPU. The input images are randomly cropped to $224 \times 224$. Moreover, we use the Adam optimizer with $\beta_1=0.9$ and $\beta_2=0.999$. The learning rate is set to $10^{-4}$ and reduced by a factor of $0.1$ for every 5 epochs. The training takes approximately 7 days.

\subsection{Aperture Disparity Map}
\label{sec4:investigation_of_aperture_flow}
\begin{figure}[t]
\centering
\includegraphics[width=1.\columnwidth]{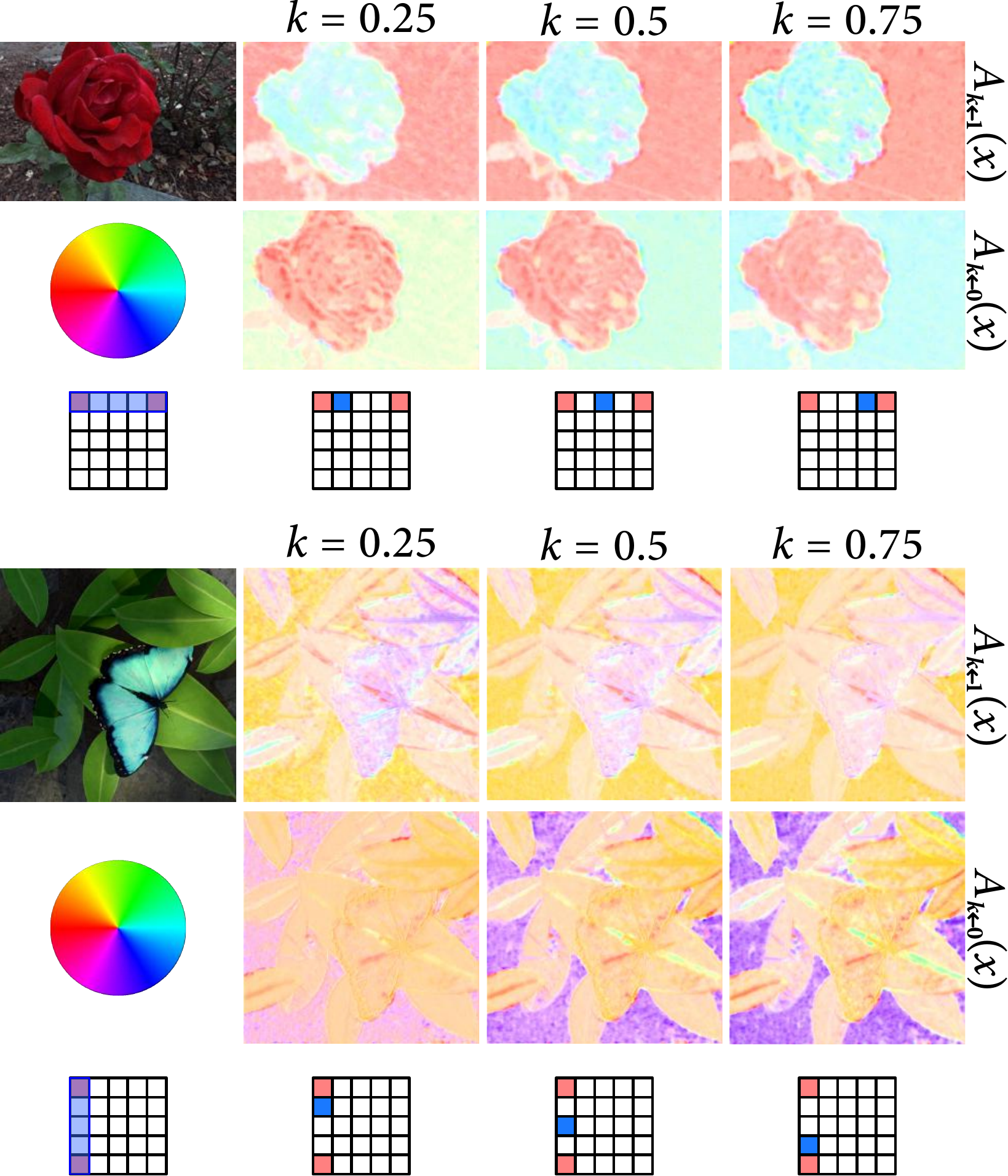}
\caption{Illustration of the ADMs from boundary images (denoted by red squares) to the target image (denoted by blue square). Actual ADMs are visualized using the color coding shown in the subfigure in the bottom-left corner of each scene. Colors represent the directions of vectors, and lighter colors mean smaller vectors.
}\label{fig:aperture_flow_vis}
\end{figure}
\begin{figure*}
	\centering
    \includegraphics[width=.98\textwidth]{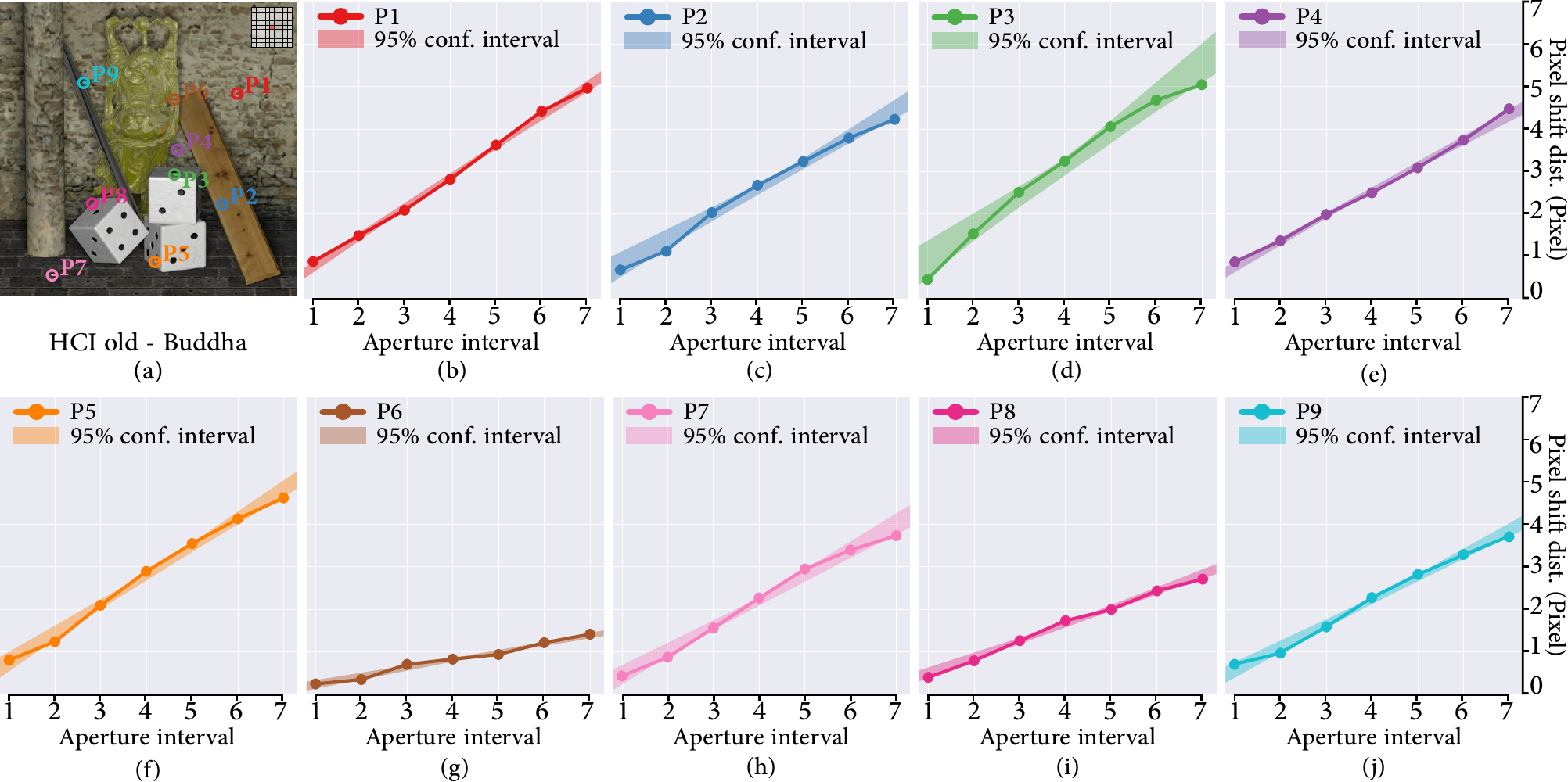}
    \caption{Relationships between the pixel shift distance and the aperture interval.}\label{fig:apertureflowlinearity}
\end{figure*}
One crucial component of our synthesis algorithm is the ADM. Each is estimated in an unsupervised manner, where the signals from the labels are directly used to reconstruct the images. \figref{fig:aperture_flow_vis} gives an example of the estimated disparity maps between the boundary images ($\mathcal{L}_0(\boldsymbol{x})$ and $\mathcal{L}_1(\boldsymbol{x})$) and different intermediate images.
We use the center $5 \times 5$ views to visualize the estimated intermediate ADMs between the images with different intervals. The input images are indicated by red squares while the ADMs are denoted by blue squares. The figure shows the ADMs between different horizontal and vertical views from both real-world and synthetic scenes. Both estimated ADMs are learned in an unsupervised manner. According to the color-coding panel, the pixels of foreground and background are shifted in opposite directions, and the colors tend to be darker when the interval between the two views is larger.

Another issue we need to tackle is a quantitative assessment of the quality of the computed ADMs. The linear relations (listed in~\cref{equ:property1,equ:property2,equ:property3,equ:property4,equ:property5,equ:property6}) offer an approach to address this. One can evaluate the linearity of the estimated ADMs in terms of the viewpoint distance between the input SAIs. To do this, we feed a sequence of image pairs with different aperture intervals into the disparity estimation network and analyze the output ADMs $A_{1 \leftarrow 0} (x)$ and $A_{0 \leftarrow 1} (x)$.

The aperture interval controls how different the two input images are. For example, if the two SAIs are with angular coordinates $(0, 0)$ and $(0,2)$ respectively, their aperture interval  is $2$. In the experiment, we track several feature points of an input image and record the shift distance of each according to the output ADMs. These feature points are selected using  corner detection~\cite{Shi1994Good}. \figref{fig:apertureflowlinearity} shows the results of $9$ randomly picked feature points  throughout the image on both the foreground objects and the background wall. The distribution of these points in a synthetic scene (\textit{Buddha}) are shown in Fig.~\hyperref[fig:apertureflowlinearity]{7a}, while the other plots (Figs.~\hyperref[fig:apertureflowlinearity]{7b} to \hyperref[fig:apertureflowlinearity]{7j})
show the relationship between the point shift distance and the aperture interval.

To further demonstrate the linearity between these two variables, we also compute their linear regression, and plot the regression line and $95\%$ confidence intervals in the Figs.~\hyperref[fig:apertureflowlinearity]{7b} to \hyperref[fig:apertureflowlinearity]{7j}.
\tabref{table:statis_table_feature_points} presents some statistics to measure the linear correlation between the shift distance and aperture interval, including the coefficient of determination ($R^2$), adjusted $R^2$, and Pearson correlation coefficient (PCC). The average values of all  three measurements are over $0.99$, which suggests that the two variables are highly linearly related.

\begin{table}[t]
\def\arraystretch{1.2}
\centering
\caption{Measurement of the linear dependence between the feature point shift distance and the aperture interval.}
\label{table:statis_table_feature_points}
\begin{tabular}{|c|c|c|c|}
	\hline
	\multirow{2}{*}{Feature Points} & \multicolumn{3}{c|}{Statistical measurements} \\ \cline{2-4} 
	                                & $R^2$      & Adjusted $R^2$  & PCC        \\ \hline \hline
	P1                              & 0.996      & 0.996           & 0.998      \\
	P2                              & 0.991      & 0.989           & 0.996      \\
	P3                              & 0.989      & 0.986           & 0.994      \\
	P4                              & 0.998      & 0.997           & 0.999      \\
	P5                              & 0.994      & 0.993           & 0.997      \\
	P6                              & 0.986      & 0.984           & 0.993      \\
	P7                              & 0.984      & 0.981           & 0.992      \\
	P8                              & 0.994      & 0.993           & 0.997      \\
	P9                              & 0.991      & 0.990           & 0.996      \\ \hline \hline
	Average                         & 0.991      & 0.990           & 0.996      \\ \hline
\end{tabular}
\end{table}

\subsection{Warping Confidence Map}
\label{sec4:investigation_of_propagation_confidence_map}
\begin{figure}
	\begin{subfigure}{0.14\textwidth}
		\centering
		\includegraphics[width=1.\columnwidth]{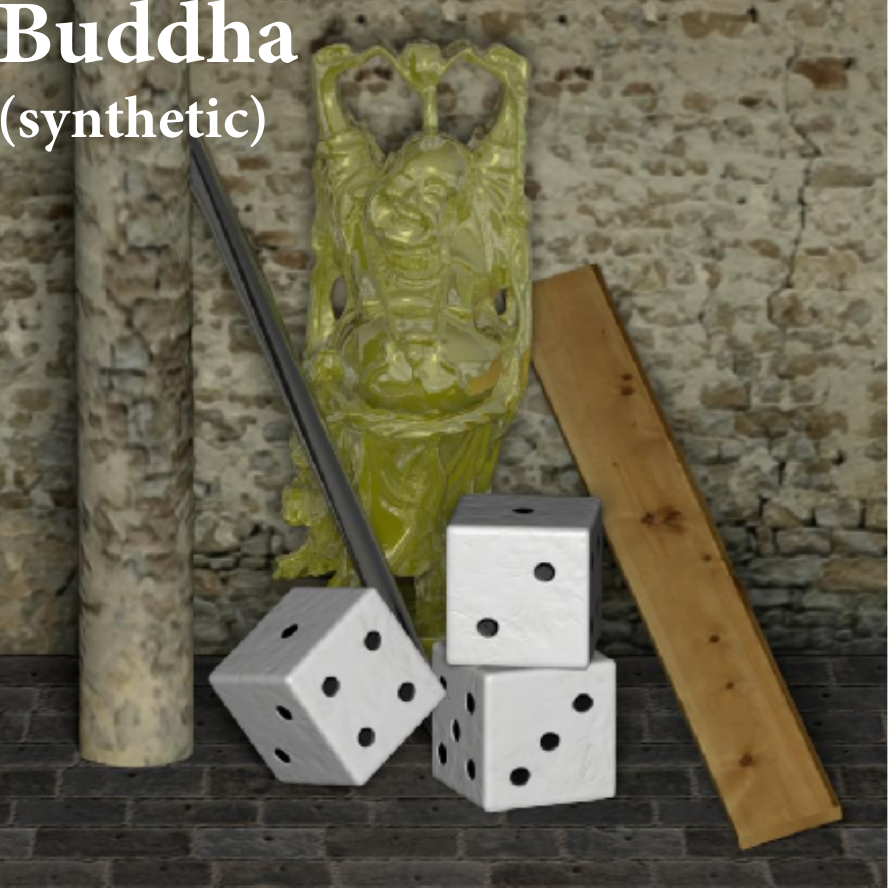} \\
		\vspace{1mm}
		\includegraphics[width=1.\columnwidth]{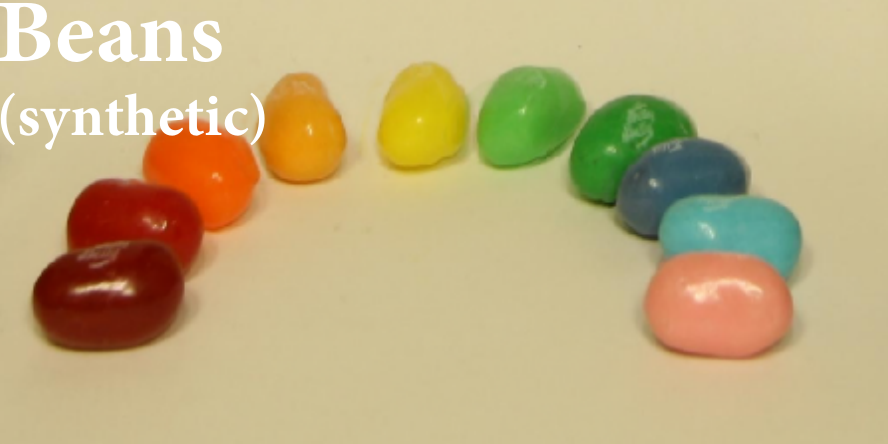} \\
		\vspace{1mm}
		\includegraphics[width=1.\columnwidth]{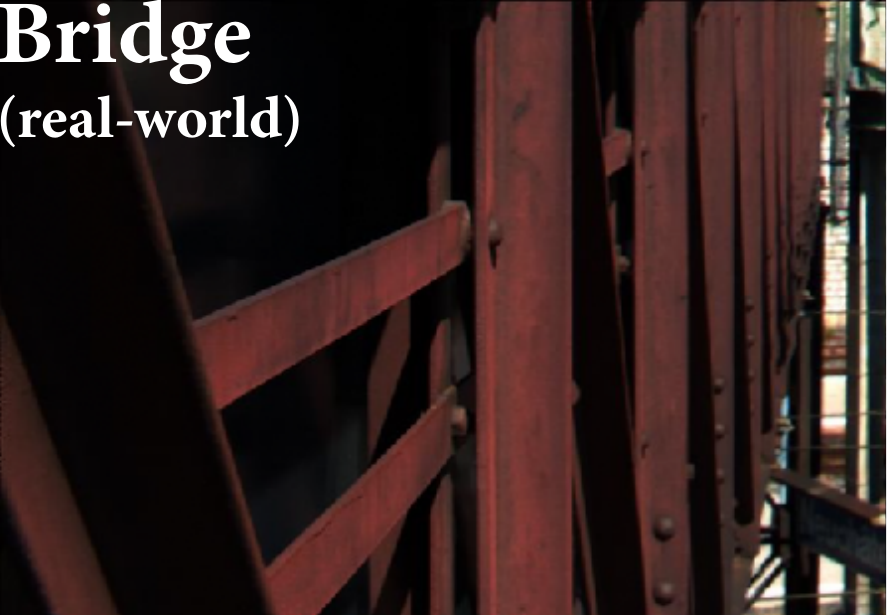} \\
		\vspace{1mm}
		\includegraphics[width=1.\columnwidth]{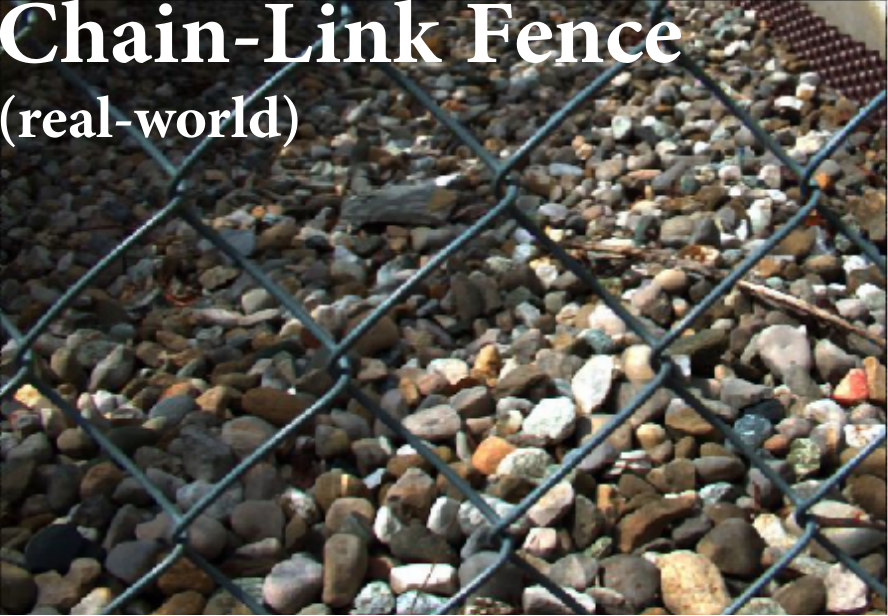}\\
		\vspace{1mm}
		\includegraphics[width=1.\columnwidth]{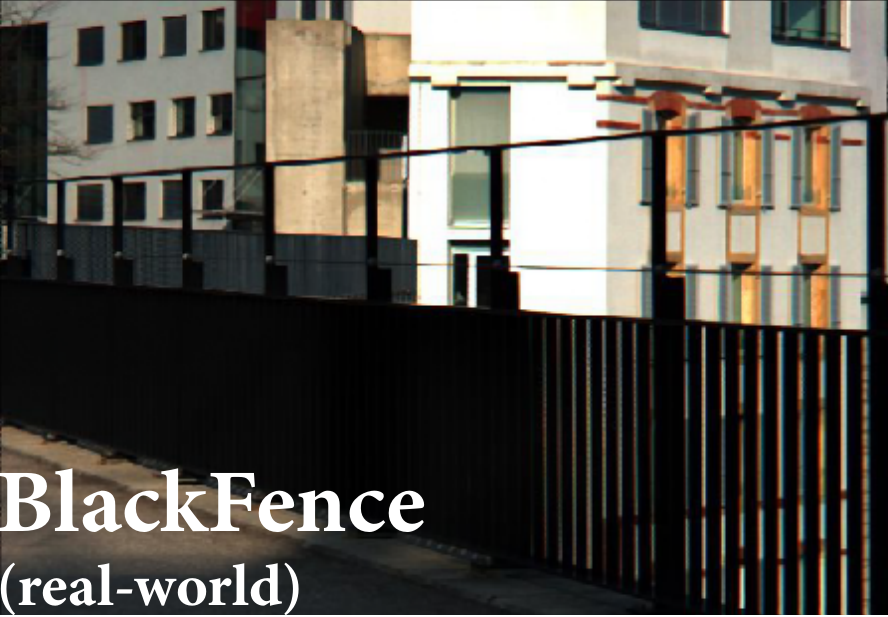}
		\caption{Scene}
	\end{subfigure}
\hspace{-1.8mm}
	\begin{subfigure}{0.14\textwidth}
	\centering
	\includegraphics[width=1.\columnwidth]{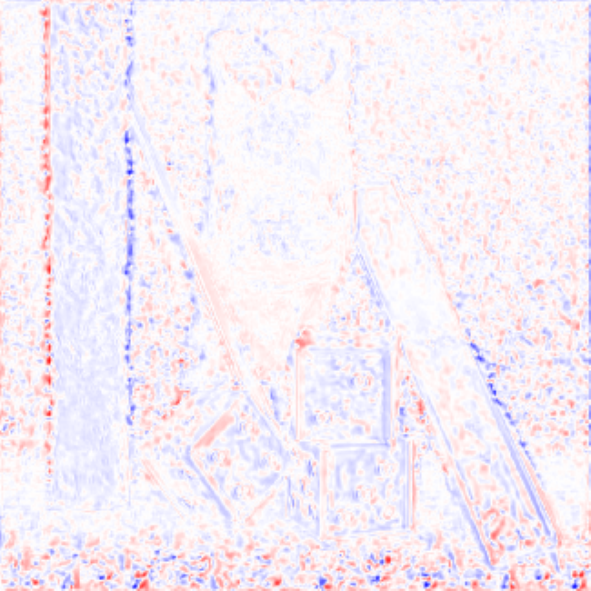} \\
	\vspace{1mm}
	\includegraphics[width=1.\columnwidth]{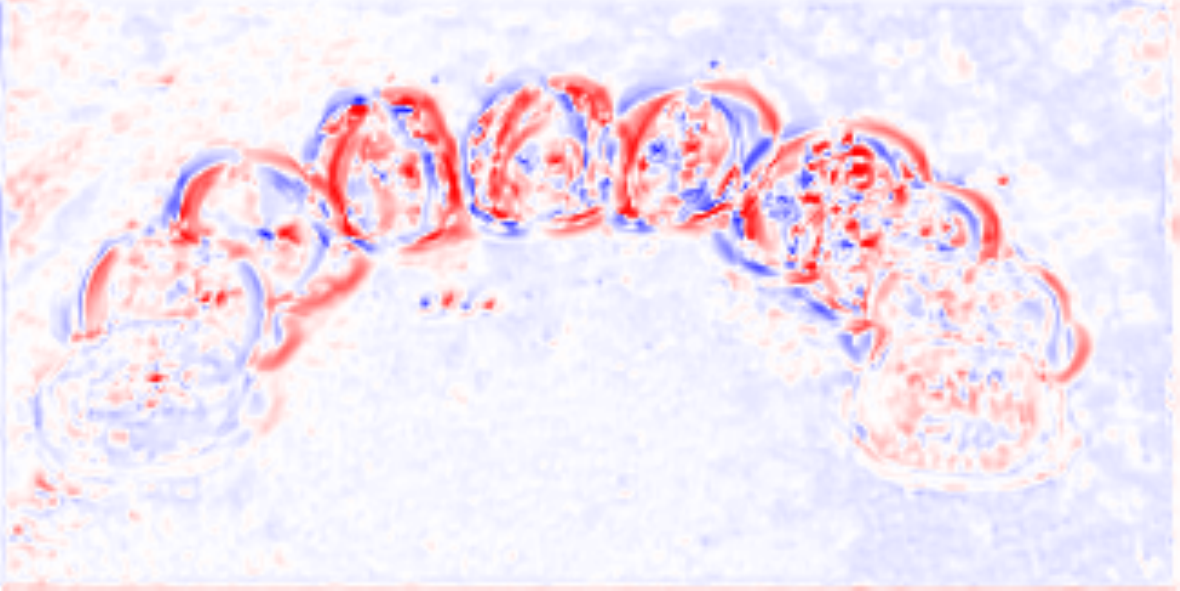} \\
	\vspace{1mm}
	\includegraphics[width=1.\columnwidth]{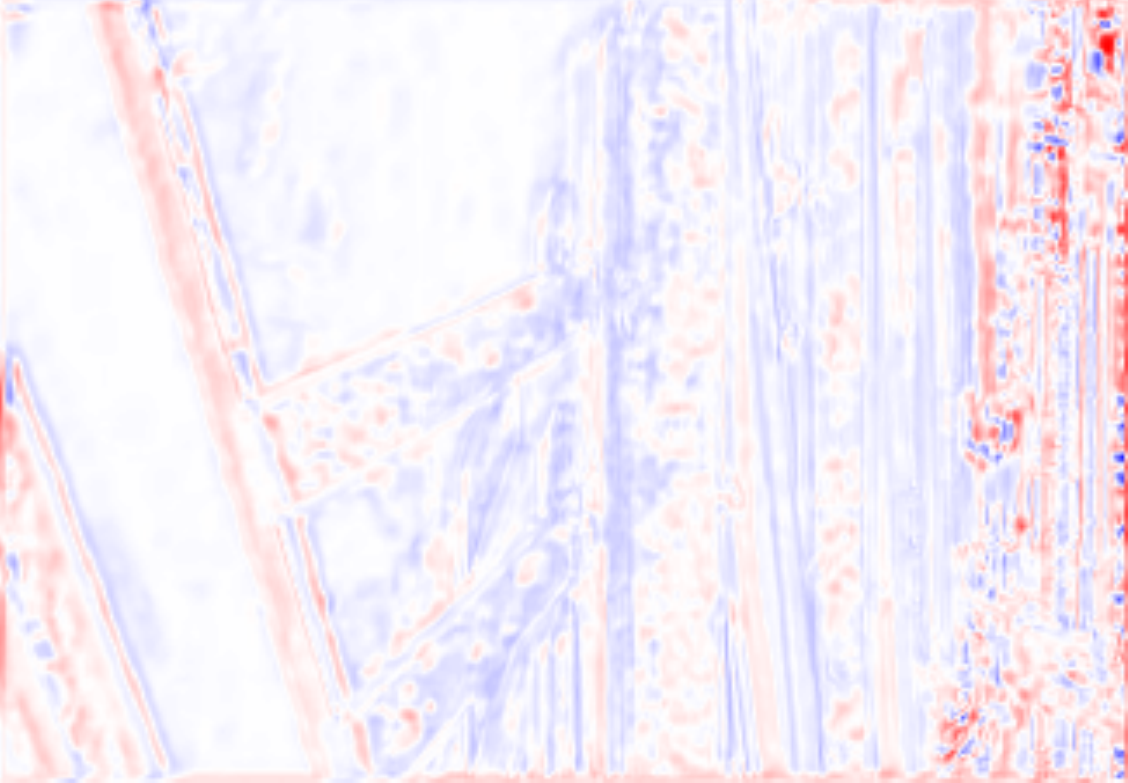} \\
	\vspace{1mm}
	\includegraphics[width=1.\columnwidth]{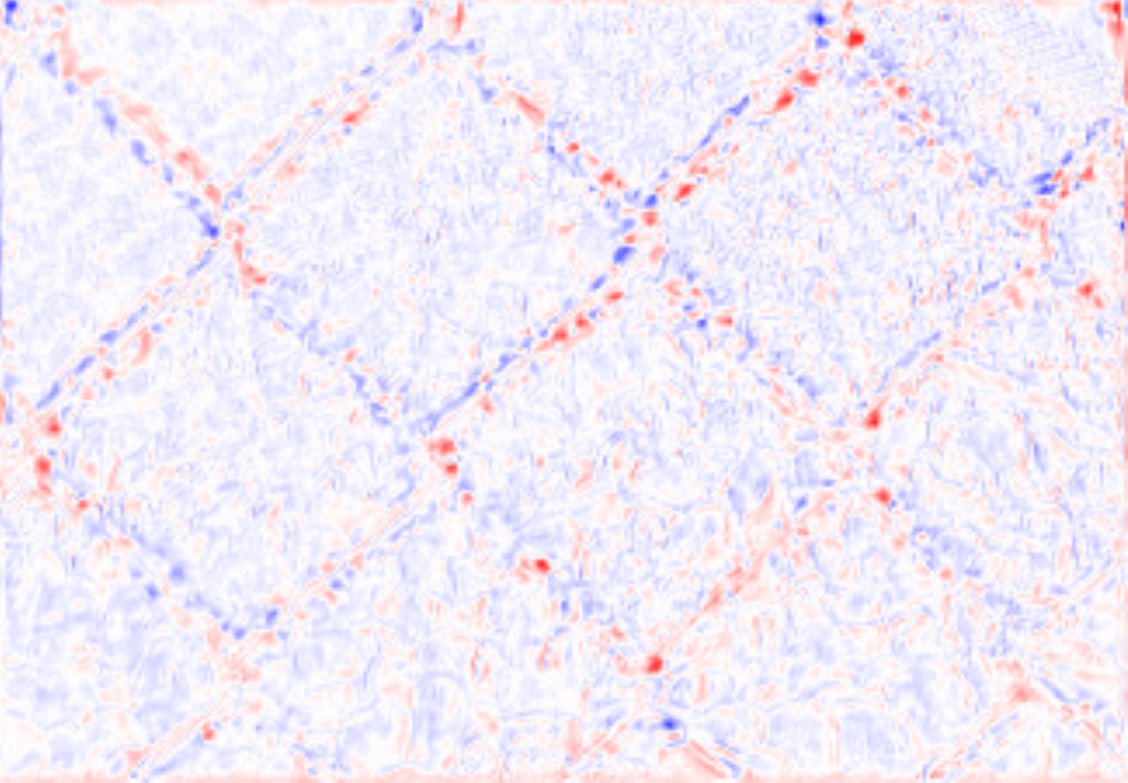}\\
	\vspace{1mm}
	\includegraphics[width=1.\columnwidth]{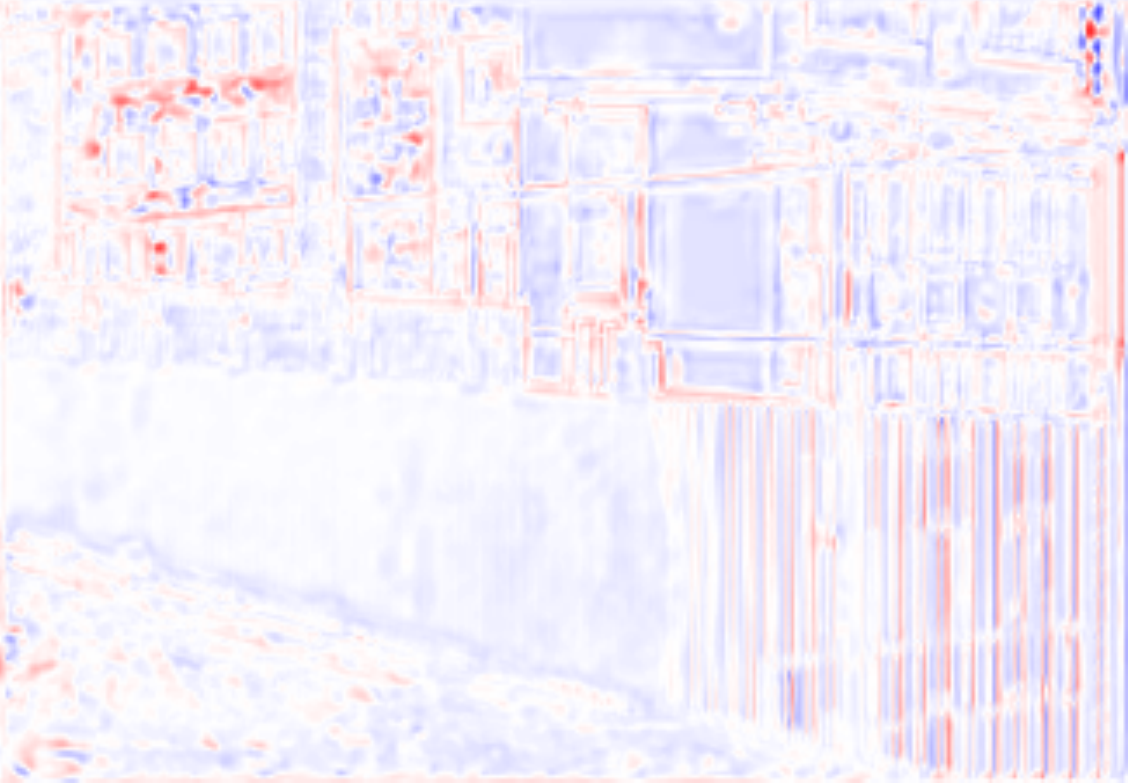}
	\caption{$O_{k \leftarrow 0}$}
	\end{subfigure}
\hspace{-1.8mm}
	\begin{subfigure}{0.185\textwidth}
	\centering
	\includegraphics[width=1.\columnwidth]{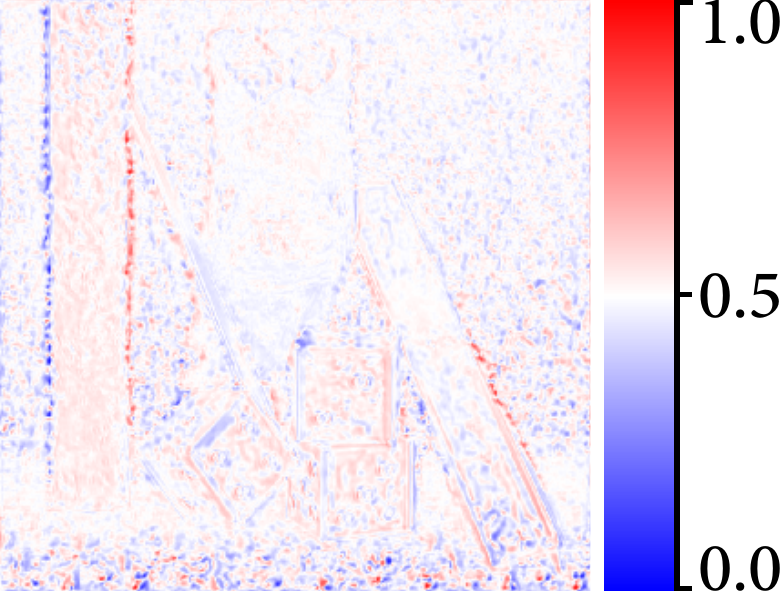} \\
	\vspace{0.8mm}
	\includegraphics[width=1.05\columnwidth]{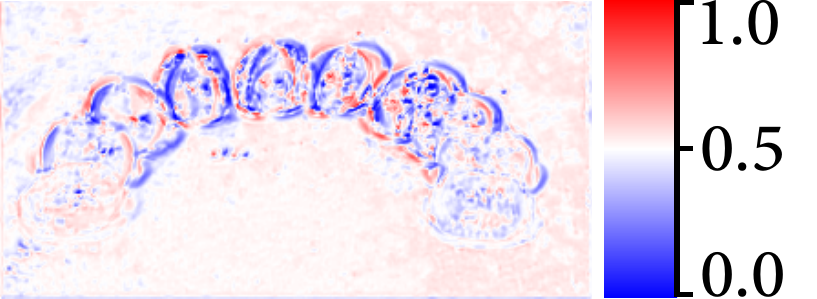} \\
	\vspace{0.8mm}
	\includegraphics[width=1.\columnwidth]{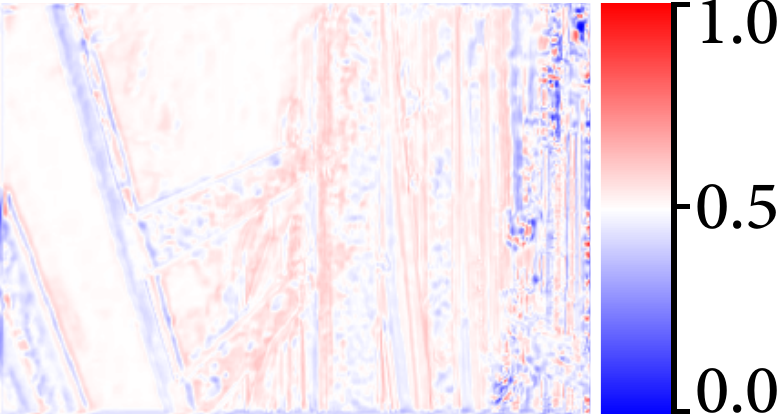} \\
	\vspace{0.8mm}
	\includegraphics[width=1.\columnwidth]{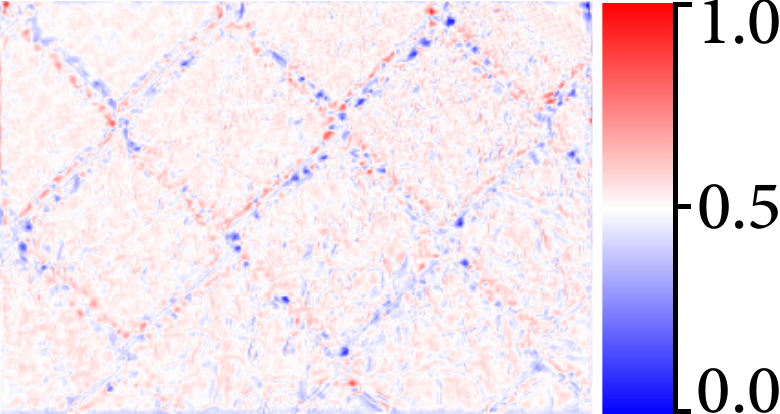}\\
	\vspace{0.8mm}
	\includegraphics[width=1.\columnwidth]{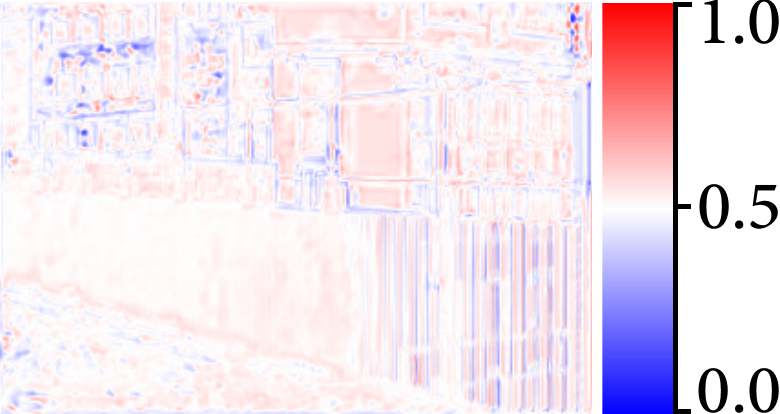}
	\caption{$O_{k \leftarrow 1}$}
	\end{subfigure}
\caption{Illustration of the WCMs of different scenes. All WCMs have $k=0.5$, i.e., they correspond to the image midway between two input images, taken from various datasets.}\label{fig:pcms}
\end{figure}

The ADM aims at estimating the displacement of each pixel in an image, but it can fail in regions with occlusions. To handle this, we introduce the WCM in order to consolidate the information from multiple views to predict a pixel value. \figref{fig:pcms} presents the WCMs of images midway between two input images, using both synthetic and real-world scenes. For each WCM, pixels in red denote those at these positions, the source image contributes more; the darker color in red, the higher contribution it makes. The case is opposite for blue.
For both real-world and synthetic scenes, the object boundary regions tend to be darker in red or blue. As discussed in Section~\ref{sec3:intermediate_radiance_inference_with_occlusions}, the occlusions always appear at the boundary regions, which magnify the distinction of the contributions (on the occluded pixels) from two input images.
Take the ``buddha'' scene (the first row in \figref{fig:pcms}) as an example. In WCMs $O_{k \leftarrow 0}$ and $O_{k \leftarrow 1}$, there are different colors near the left and right boundaries of the pillar. This demonstrates that the left input image ($\mathcal{L}_0$) contributes more on the left boundary region while the right input image ($\mathcal{L}_1$) contributes more on the right boundary region, as interpreted in \figref{fig:occlusions_analysis} (Line $P'MO''$ and Line $R'NQ''$).

	\section{Results and Discussions}
	\label{sec:results}
	To validate the effectiveness of our proposed framework, we conduct experiments on both synthetic and real-world light fields. We use common classical quantitative metrics, namely peak signal-to-noise ratio (PSNR) and structural similarity index (SSIM) to assess the performance of the algorithms.

\subsection{Comparison with Continuous Synthesis Methods}
\label{sec5:subsec:comparision_with_continuous_synthesis_methods}

\begin{figure*}
	\centering
	\includegraphics[width=1.\textwidth]{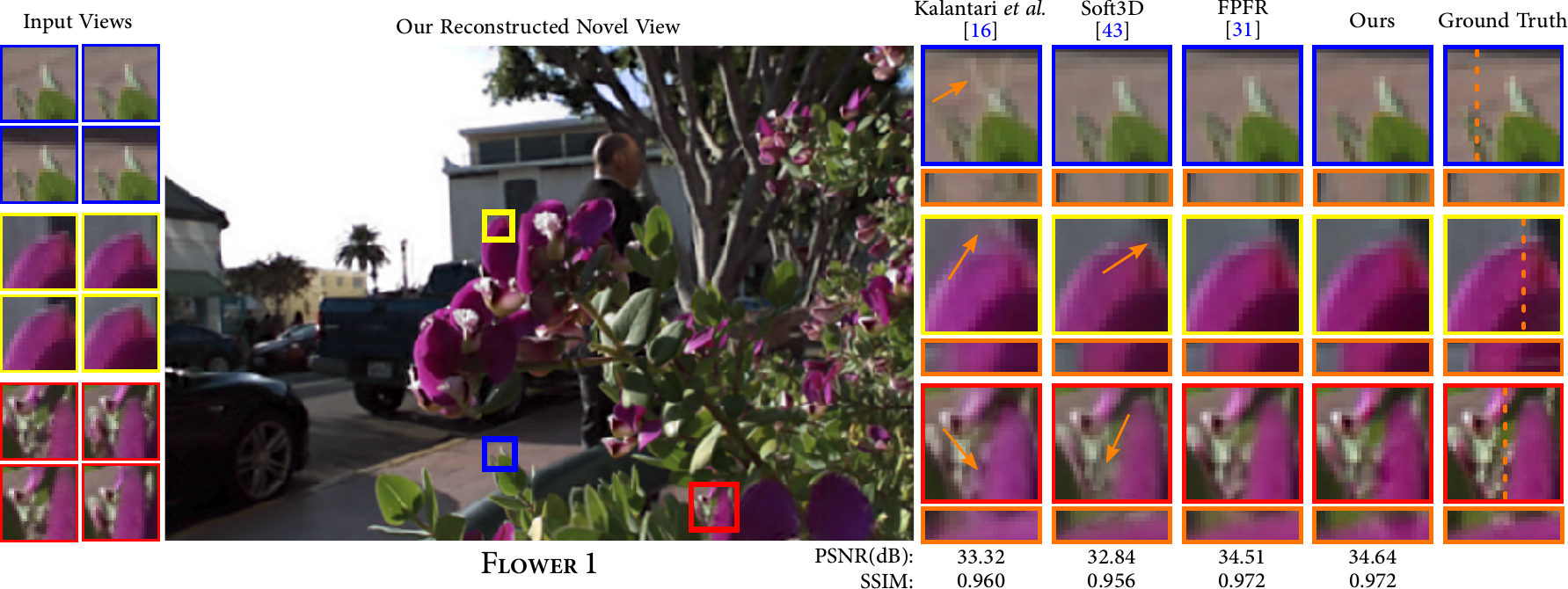} \newline \vspace{-0.5em}
	\includegraphics[width=1.\textwidth]{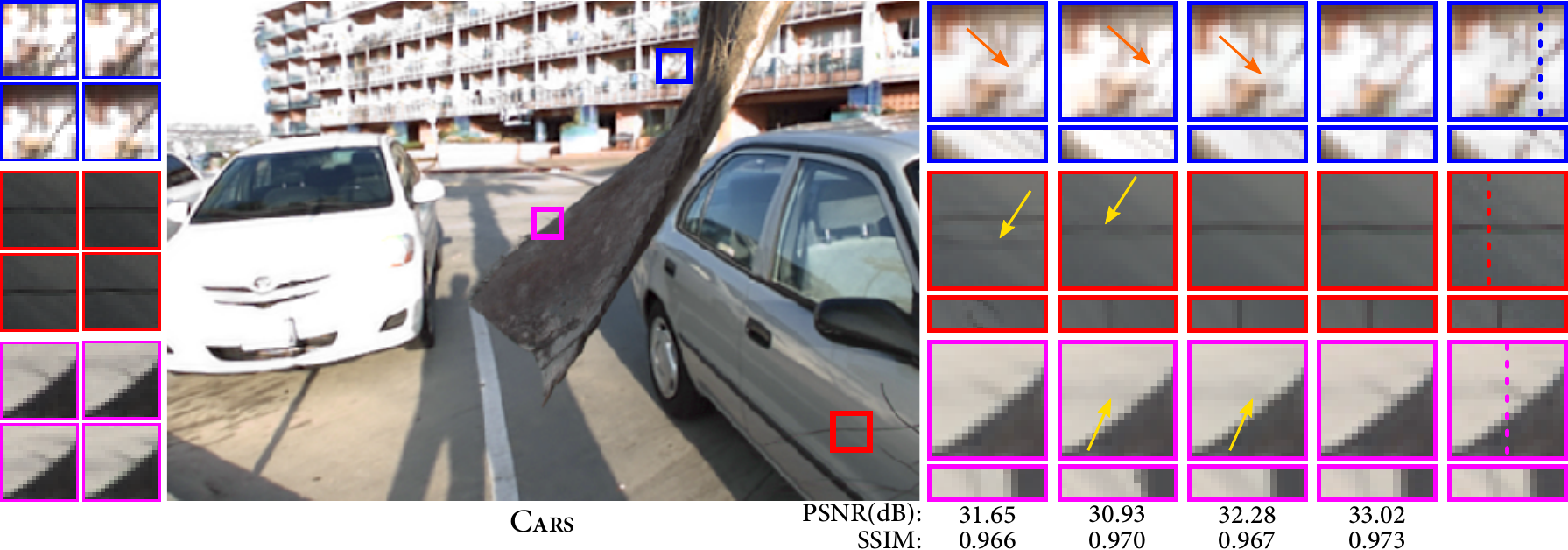}
	\caption{We compare our approach against multiple state-of-the-art methods on several difficult Lytro light field scenes from \etal{Kalantari}~\cite{Kalantari2016Learning}. The scene ``Flower 1'' contains a significant number of occluded regions and the scene ``Cars'' contains many thin structures, such as the fibre of the paper (in the magenta box) and the reflection of tree branches (in the red box). Our approach produces more realistic results compared with the selected algorithms. We also show the EPI at the dashed line in each zoomed region.}
	\label{fig:continuous_reconstruction}
\end{figure*}

\begin{table}[t]
	\centering
	\setlength{\tabcolsep}{2pt}
	\renewcommand{\arraystretch}{1.5}
	\caption{Quantitative comparisons (PSNR/SSIM) of the proposed approach with continuous view synthesis algorithms for angular super-resolution $2\times2 \rightarrow 8\times8$. The input light fields for each algorithm are sampled at four corners.}\label{table:continues_algorithms}
	\begin{minipage}{\columnwidth}
	\begin{tabular}{|c|cccc|}
		\hline
		\multicolumn{1}{|c|}{\multirow{2}{*}{Scenes}} & \multicolumn{4}{c|}{Methods}                                                                                        \\ \cline{2-5} 
		\multicolumn{1}{|c|}{}                        & \multicolumn{1}{c|}{\etal{Kalantari}~\cite{Kalantari2016Learning}}  & \multicolumn{1}{c|}{Soft3D~\cite{Penner2017Soft}\footnotemark{}} & \multicolumn{1}{c|}{FPFR~\cite{Shi2020Learning}} & \textbf{Ours}  \\ \hline \hline
		Flowers1    & 33.32 / 0.960 & 32.84 / 0.960  & 34.51 / \textbf{0.972}          & \textbf{34.64} / \textbf{0.972} \\
		Flowers2    & 31.94 / 0.960 & 32.74 / 0.961  & \textbf{34.20} / \textbf{0.972} & 34.03 / 0.965 \\
		Cars        & 31.65 / 0.966 & 30.93 / 0.962  & 32.28 / 0.967                   & \textbf{33.02} / \textbf{0.973} \\
		Seahorse    & 31.87 / 0.970 & 31.22 / 0.961  & \textbf{34.36} / 0.970          & 33.61 / \textbf{0.972} \\
		StoneLion   & 40.57 / 0.979 & 38.53 / 0.973  & 40.02 / 0.975                   & \textbf{40.84} / \textbf{0.981} \\ 
		Leaves1     & 35.84 / 0.973 & 31.74 / 0.947  & 34.76 / 0.966                   & \textbf{36.95} / \textbf{0.978} \\ 
		Leaves2     & 34.17 / 0.963 & 32.47 / 0.965  & 32.86 / 0.957                   & \textbf{34.28} / \textbf{0.966} \\ \hline 
		Average     & 34.19 / 0.967 & 32.92 / 0.961  & 34.71 / 0.969                   & \textbf{35.33} / \textbf{0.972} \\ \hline		
	\end{tabular}
	\footnotetext{\blue{*}Note that Soft3D~\cite{Penner2017Soft} and FPFR~\cite{Shi2020Learning} did not release their codes. For FPFR, we used the code provided by the authors. For Soft3D, we used the code reimplemented by the authors of reference~\cite{Jiang2019Learning}.}
	\vspace{-1em}
	\end{minipage}
\end{table}
We first evaluate the performance of our method against the recent techniques for continuous view synthesis. Algorithms of this kind attempt to learn a continuous representation of the plenoptic function from a sparsely-sampled input light field. The comparisons are conducted against three recent learning-based methods, namely \etal{Kalantari}~\cite{Kalantari2016Learning}, Soft3D~\cite{Penner2017Soft} and
\etal{Shi}~\cite{Shi2020Learning}. Table~\ref{table:continues_algorithms} presents the quantitative results of different algorithms on the Lytro images for angular $2\times2 \rightarrow 8\times8$ resolution enhancement. As shown, our method achieves the best quantitative results compared with all these methods.

\figref{fig:continuous_reconstruction} compares the visual performance of different algorithms. As shown, the approach of \etal{Kalantari} tends to generate artifacts near the object boundary, such as the leaves and the petal boundary in ``Flower 1''. Soft3D~\cite{Penner2017Soft} gives relatively smooth results near the petal boundary. As for ``Cars'', all of the other three methods lose the information of the thin structures. In this picture, three regions (the colored boxes) with thin structures are selected and zoomed in to highlight the differences. For each zoomed region, we also show the EPI near the dashed line. In comparison, our model produces more realistic spatial results for both thin objects and boundary regions. The disparity is hard to estimated near the thin structures, leading to the missing or aliasing of these structures in the reconstructed EPIs. From the EPI results in the scene ``Cars'', one can also see that the reconstructed light field using our method also retain better disparity information. In both examples, our results are closer to the ground truth. In addition, our approach takes about 0.23s per synthesized view on average to reconstruct a $8\times8$ light field from its $2\times2$ views at a spatial resolution of $376 \times 541$, which is nearly $40$ times faster than \etal{Kalantari} (about $9$s per view) and $3.7$ times faster than FPFR~\cite{Shi2020Learning} (about $0.85$s per view).

\begin{figure*}
	\centering
	\includegraphics[width=1.\textwidth]{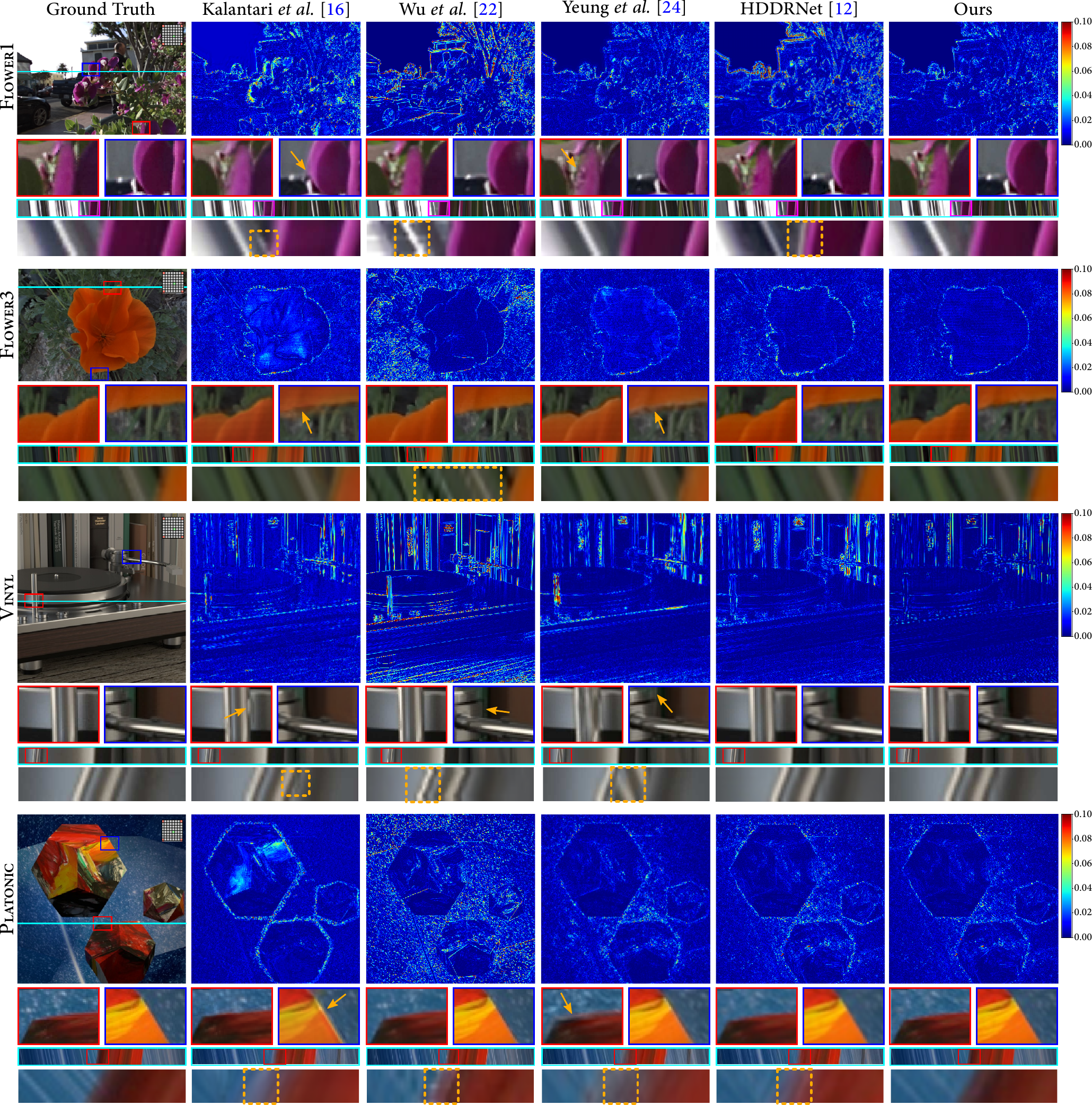}
	\caption{Visual comparisons of different algorithms on the $(5, 5)$ synthesized view for the task $2\times2 \rightarrow 8\times8$ on both real-world and synthetic scenes. We select two regions (red and blue boxes) to compare the spatial results of different algorithms. For each reconstructed light field, the EPIs at the position highlighted by the cyan line are also visualized. We zoom in one selected region (red box) of each EPI for better comparison, and in the zoomed EPI we highlight the region that is obviously different from the ground truth.}
	\label{fig:2x2_8x8_learning_based}
\end{figure*}

\begin{table*}[t]
	\centering
	\setlength{\tabcolsep}{6pt}
	\renewcommand{\arraystretch}{1.5}
	\caption{Quantitative comparisons (PSNR/SSIM) of the proposed approach with learning-based methods.}
	\label{table:learning_based_algorithms}
	\begin{tabular}{|c|c|ccccc|}
		\hline
		\multicolumn{1}{|c|}{\multirow{2}{*}{Scenes}} & \multicolumn{1}{|c|}{\multirow{2}{*}{Data Type}} & \multicolumn{5}{c|}{Methods}             \\ \cline{3-7} 
		\multicolumn{1}{|c|}{}                        & \multicolumn{1}{|c|}{}   & \multicolumn{1}{c|}{\etal{Kalantari}~\cite{Kalantari2016Learning}} & \multicolumn{1}{c|}{\etal{Wu}~\cite{Wu2018Light}} & \multicolumn{1}{c|}{\etal{Yeung} ~\cite{Yeung2018Fast}} & \multicolumn{1}{c|}{HDDRNet~\cite{Meng2019Highdimensional}} & \textbf{Ours}   \\ \hline \hline
		Bedroom         &  Synthetic & 34.87 / 0.914 & 32.52 / 0.890 & 34.26 / 0.895 & 34.29 / 0.892 & \textbf{34.48} / \textbf{0.906} \\
		Cotton          &  Synthetic & 41.98 / 0.964 & 38.49 / 0.944 & 41.47 / 0.956 & 42.32 / 0.961 & \textbf{43.63} / \textbf{0.973} \\
		Dino            &  Synthetic & 36.88 / 0.951 & 34.21 / 0.904 & 40.38 / 0.951 & 40.34 / 0.952 & \textbf{40.78} / \textbf{0.957} \\
		Origami         &  Synthetic & 30.91 / 0.903 & 28.26 / 0.898 & 31.54 / 0.912 & 32.18 / 0.912 & \textbf{33.59} / \textbf{0.919} \\ \hline
		Average         &  ---       & 36.16 / 0.933 & 33.37 / 0.909 & 36.91 / 0.929 & 37.28 / 0.929 & \textbf{38.12} / \textbf{0.939} \\ \hline \hline
		Occlusions      & Real-world & 32.18 / 0.897 & 30.48 / 0.867 & \textbf{33.19} / 0.908 & 32.78 / 0.909 & 33.10 / \textbf{0.912} \\
		Reflective	    & Real-world & 35.28 / 0.923 & 33.21 / 0.893 & 36.82 / 0.933 & 36.77 / 0.931 & \textbf{37.01} / \textbf{0.950} \\ \hline
		Average         &  ---       & 33.73 / 0.910 & 31.85 / 0.880 & 35.01 / 0.920 & 34.78 / 0.920 & \textbf{35.06} / \textbf{0.931} \\ \hline 
	\end{tabular}
\end{table*}

\subsection{Comparison with Other Learning-Based Methods}
\label{sec5:subsec:comparison_with_learning_based_methods}
Next, we compare the performance of our method against several recent learning-based methods, including \etal{Wu}~\cite{Wu2018Light}, \etal{Yeung}~\cite{Yeung2018Fast} and HDDRNet~\cite{Meng2019Highdimensional}. Since the method of \etal{Kalantari}~\cite{Kalantari2016Learning} has been compared with these methods in the respective paper, we also include it as a reference. Table~\ref{table:learning_based_algorithms} and \figref{fig:2x2_8x8_learning_based} present the quantitative and visual results, respectively.
\etal{Wu}~\cite{Wu2018Light} design a ``blur-restoration-deblur'' pipeline to overcome angular aliasing, but the algorithm requires at least three views to generate acceptable results. For $2\times2 \rightarrow 8\times8$ task, because only two views are available for inputs, the insufficient angular information leads to aliasing effects in their reconstructed EPIs, especially in the EPIs of the two Flower scenes in \figref{fig:2x2_8x8_learning_based}.
\etal{Yeung}~\cite{Yeung2018Fast} and HDDRNet~\cite{Meng2019Highdimensional} achieve state-of-the-art performance. They both adopt the 4D convolution to fully exploit the spatio-angular information, which result in their superior results. 

However, for the synthetic scenes, their results tend to have more errors near the object boundaries.
Another drawback for these end-to-end methods is that they can only generate the light fields with fixed angular dimensions. This is due to the reshape operation (between the channel and angular dimensions) when upsampling the angular resolution.
As a result, when the number of input (or output) views varies, both have to train a new model to fit for such changes. In comparison, we increase the number of views by utilizing the continuous representation among the views and adopt the fully-convolutional framework as the refinement module. Such design enables our approach to handle the input light field with various angular dimension, and the resulting approach achieves the best quantitative results and can produce images close to the ground truth.

\subsection{Ablation Study}
\label{sec5:subsec:ablation_study}
\begin{figure}[t]
    \centering
    \includegraphics[width=1.\columnwidth]{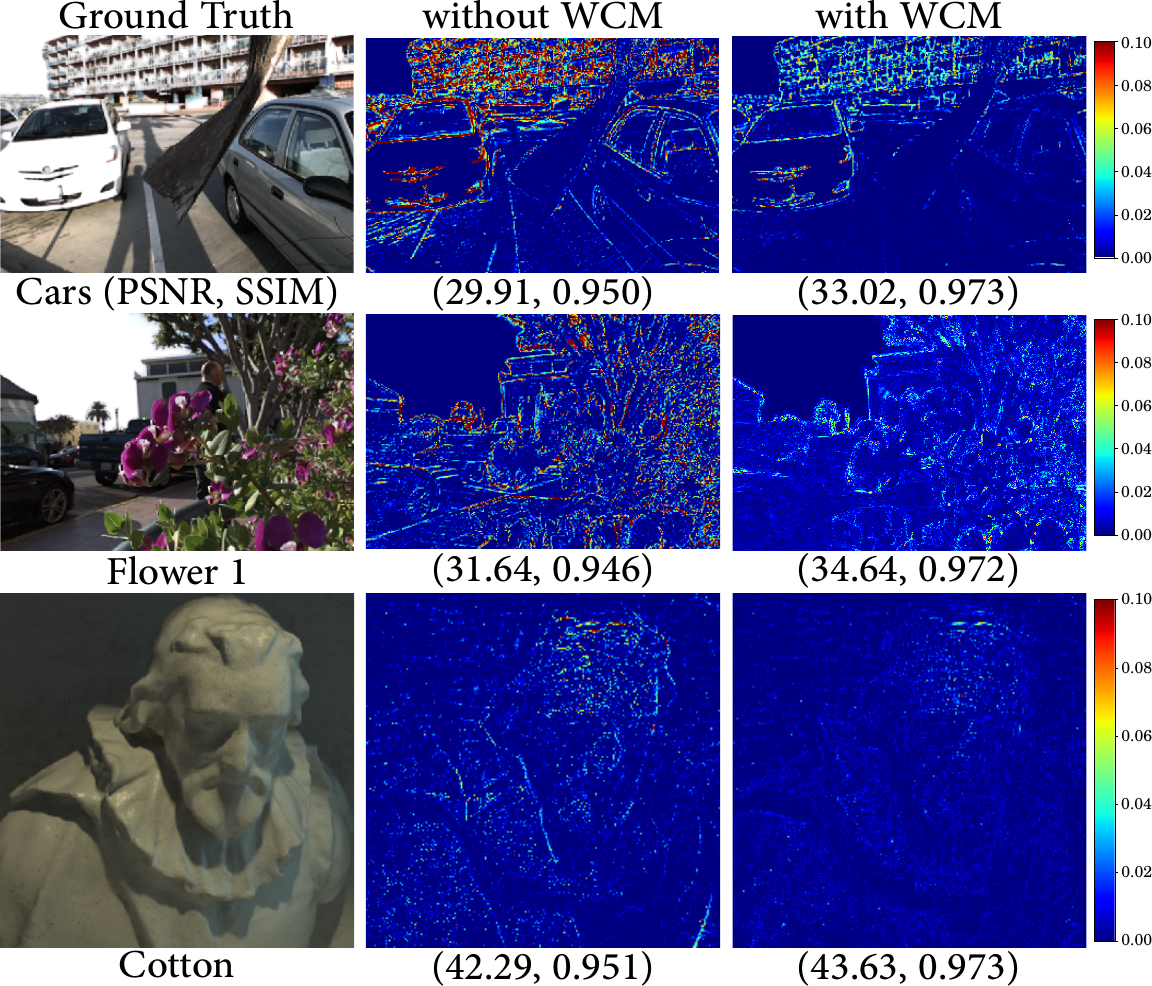}
    \caption{Performance comparison of the proposed model with and without WCM module.}
    \label{fig:abstudy_WCM}
\end{figure}

The ablation studies are conducted on several crucial components to evaluate our framework. Regarding WCM, we compare the performance of the variants with and without this module, and present the results in Fig.~\ref{fig:abstudy_WCM}.
Comparing the error maps in the second and third columns, we can observe an obvious drop on the accuracy of reconstruction near the object boundary regions if WCM is removed. For the scenes contain complex occlusions, such as ``Cars'' and ``Flower 1'', the quantitative measurement PSNR reduces even more, by about 3dB.
\begin{figure}[t]
    \centering
    \includegraphics[width=1.\columnwidth]{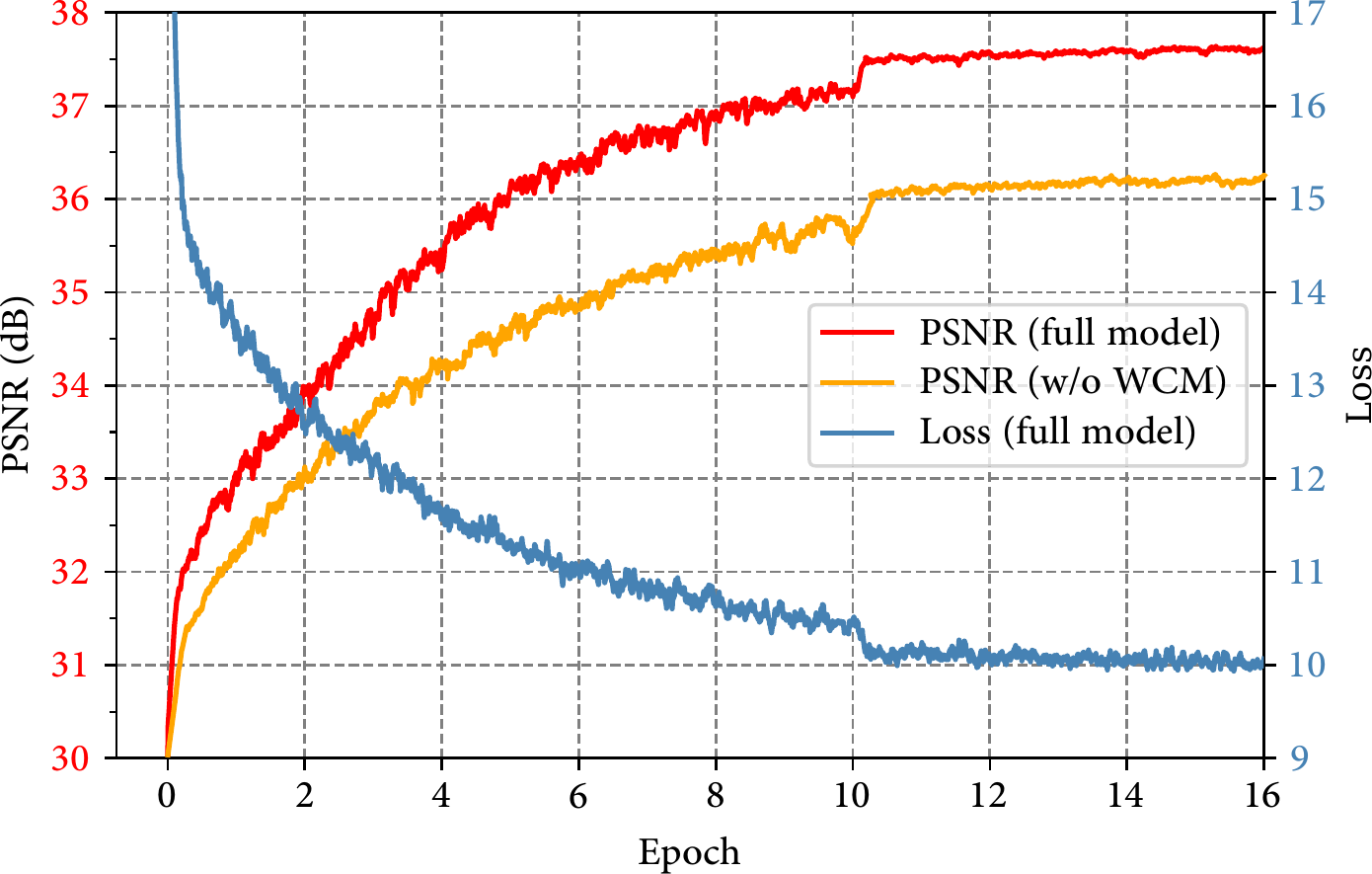}
    \caption{Learning curves of different variants of the proposed model.}
    \label{fig:convergence_behavior}
\end{figure}
In addition to the results, we also explore the convergence behavior of different variants, as shown in Fig.~\ref{fig:convergence_behavior}. By properly handling the occlusions, our model can converge to a better point.

\begin{figure}[t]
	\centering
	\includegraphics[width=1.\columnwidth]{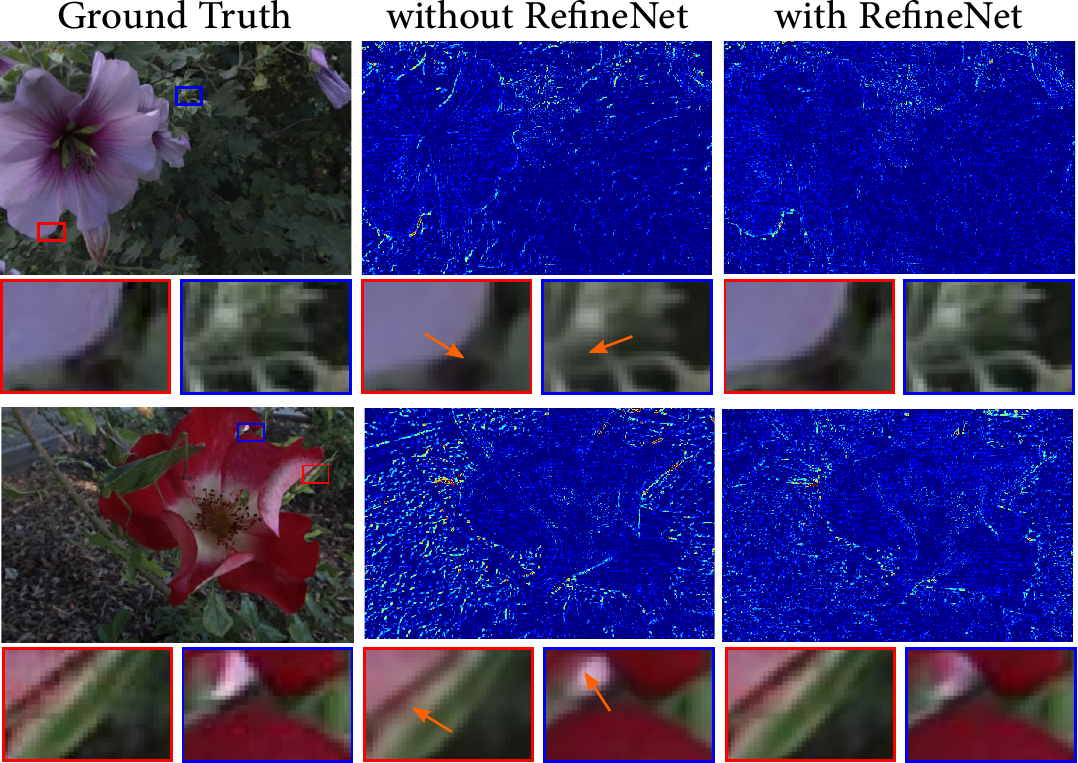}
	\caption{Illustration of the visual improvements by adding the RefineNet module. The second and third columns present the reconstruction error maps without and with RefineNet.
	}\label{fig:ablation_refine_visual_comparison}
\end{figure}

Second, regarding RefineNet, we compare the light fields generated by our approach with and without this module. Table~\ref{table:refinenet_evaluation} shows the quantitative results on both real-world and synthetic scenes. We use 20 real-world scenes randomly selected from the ``Flower'' dataset~\cite{Srinivasan2017Learning}. Each light field contains a flower in the foreground, which has a relatively larger disparity than the background. The synthetic scenes are randomly selected from HCI~\cite{Honauer2016Dataset} and we use 10 light fields. As shown, the RefineNet module improves the reconstruction results by over 1.3dB on PSNR, and it also shows improvement in terms of SSIM. \figref{fig:ablation_refine_visual_comparison} illustrates the visual improvements by adding this module. The reconstructed errors are further reduced and some spatial details can be recovered better.

Third, our ablation studies also evaluate the functions of different loss terms. The reconstruction loss term $\ell_r$ is essential for the network training, and therefore we conduct the ablation studies to evaluate the effectiveness of the other three terms. Table~\ref{table:ablation_quantitative_flower20} shows the quantitative results of the proposed network (without the RefineNet module) trained with different loss terms on the ``Flower'' dataset. As shown, the full model achieves the best performance.

\begin{table}[t]
	\centering
	\setlength{\tabcolsep}{6pt}
	\renewcommand{\arraystretch}{1.36}
	\caption{Verifying the effectiveness of RefineNet. We compare the reconstruction quality of the light fields generated with and without this module under the task $2\times2 \rightarrow 8\times8$ over 20 Flower scenes~\cite{Srinivasan2017Learning} and 10 HCI scenes~\cite{Honauer2016Dataset}.}\label{table:refinenet_evaluation}
	\begin{tabular}{|c|cccc|}
		\hline
		\multicolumn{1}{|c|}{\multirow{2}{*}{Metrics}} & \multicolumn{2}{c|}{without RefineNet}     & \multicolumn{2}{c|}{with RefineNet}       \\\cline{2-5} 
		\multicolumn{1}{|c|}{}                         & Flower (20) & \multicolumn{1}{c|}{HCI (10)} & Flower (20) & HCI (10)  \\ \hline \hline
		PSNR                                           &    35.33     &     33.56         &    37.05    &    34.92    \\
		SSIM                                           &    0.945     &     0.912         &    0.956    &    0.930    \\
		\hline
	\end{tabular}
\end{table}
\begin{table}[t]
	\centering
	\setlength{\tabcolsep}{6pt}
	\renewcommand{\arraystretch}{1.36}
	\caption{Ablation study on different loss terms for training the proposed framework on the Flower dataset~\cite{Srinivasan2017Learning}.}\label{table:ablation_quantitative_flower20}
	\begin{tabular}{r|cc}
		\hline
		\multicolumn{1}{c|}{Settings}      & PSNR                 & SSIM                 \\ \hline
		without perceptual loss            & 35.32                & 0.943                \\
		without warping loss               & 35.29                & 0.920                \\
		without smoothness loss            & 35.33                & 0.944                \\ \hline
		Full loss                          & \textbf{35.33}       & \textbf{0.945}       \\ \hline
	\end{tabular}
\end{table}

\subsection{Application on Image-Based Rendering}
\label{sec5:subsec:application}
\begin{figure*}[t]
	\centering
	\includegraphics[width=.9\textwidth]{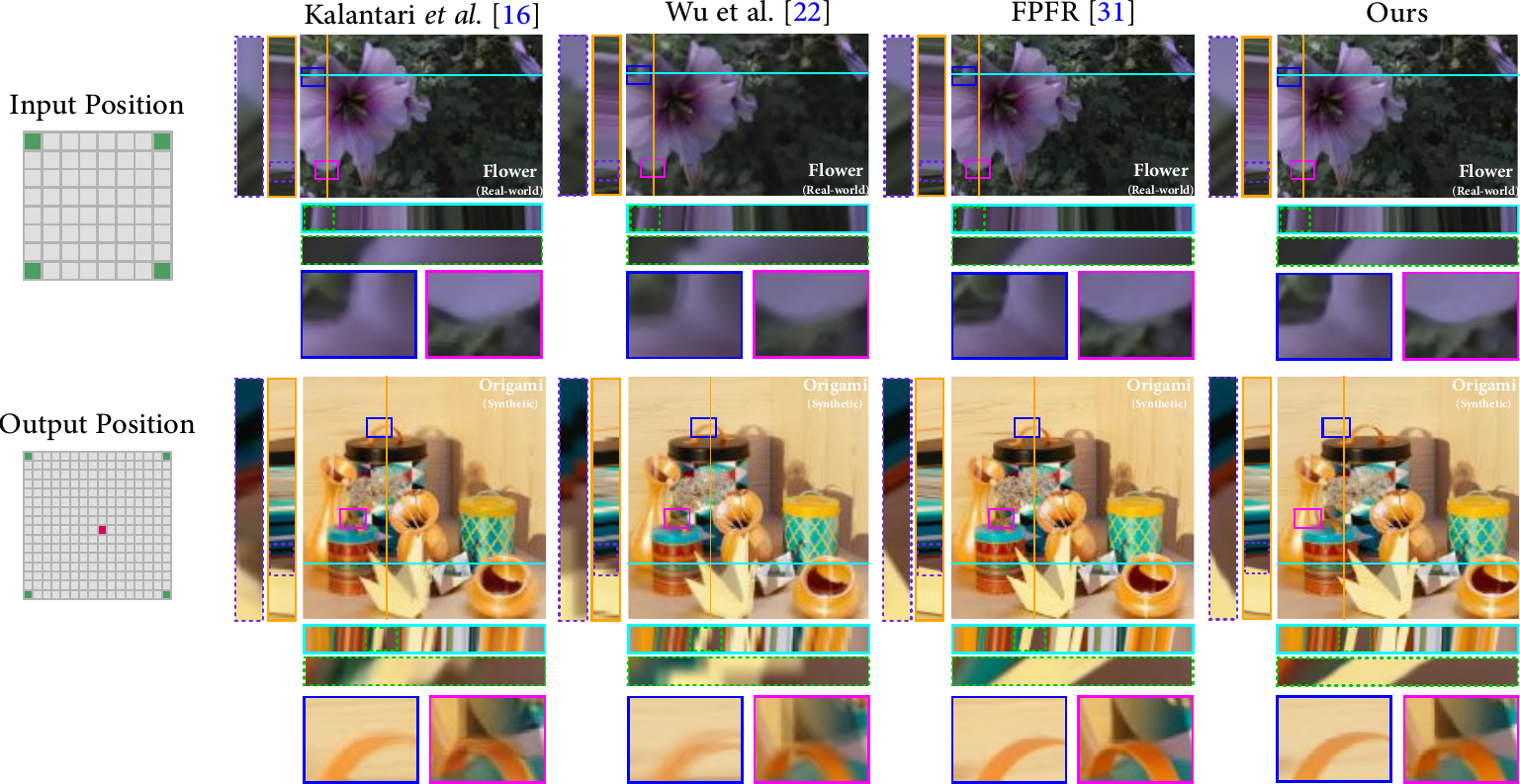}
	\caption{Visual comparisons on light field dense reconstruction on both real-world (top) and synthetic scenes (bottom). We present the results of $16\times16$ reconstruction from $2\times2$ corner views sampled from the $8\times8$ input image grid. The $(9, 9)^\mathrm{th}$ SAI (the red square) of the light field recovered from different methods are presented. Three selected regions have been zoomed in for clarity. Horizontal and vertical EPIs corresponding to the highlighted lines with different colors are shown, and a selected region in each EPI is zoomed in for better viewing.}\label{fig:arbitrary_reconstruction}
\end{figure*}
IBR aims at producing the images at new camera positions from a set of captured samples. For light field, one major benefit of rendering-based methods is that they do not necessarily require explicit geometric models and can generate the new views by straightforward interpolation~\cite{Levoy1996Light}. However, in order to produce the plausible views, such techniques often require the light field to be densely sampled~\cite{Chai2000Plenoptic}. In comparison, our method can reconstruct the densely-sampled light field with sufficient angular resolution to enable the rendering applications.

To evaluate the effectiveness of our method on image-based rendering applications, we compare the performance of dense reconstruction on both synthetic and real-world scenes. Specifically, we compare the performance of different algorithms when reconstructing $16\times16$ densely-sampled light fields from $2\times2$ corner views sampled from the $8\times8$ input image grid. \figref{fig:arbitrary_reconstruction} visualizes the $(9, 9)^\mathrm{th}$ SAI of the reconstructed light field images. As shown, compared with \etal{Kalantari}~\cite{Kalantari2016Learning} and \etal{Wu}~\cite{Wu2018Light}, our method produces more realistic results with sharp textures, and constructs the EPIs with clear slopes.
We also compare with FPFR~\cite{Shi2020Learning}, and both methods have competitive performance, according to the visual results shown in the last two columns of Fig.~\ref{fig:arbitrary_reconstruction}.


	\section{Conclusions}
	\label{sec:conclusions}
	In this paper, we propose a depth-free algorithm for the reconstruction of arbitrary intermediate views of the light fields. To efficiently describe the parallax between any two given views, we define the ADM based on the epipolar property of the light field. By incorporating also the WCM, our method can efficiently address the occlusions near the object boundaries. Both ADM and WCM are approximated using a dense network. In addition, we further adopt a 4D CNN with alternating filters in the refinement stage to improve the quality of synthesized images. Experimental results have demonstrated that the proposed model achieves state-of-the-art performance for both synthetic and real-world light fields.
	
	\section{Acknowledgments}
	\label{sec:acknowledgments}
	This work is supported in part by the Research Grants Council of Hong Kong (GRF 17201818, 17200019, 17201620) and the University of Hong Kong (104005438, 104005864). We thank Dr. Xiaoran Jiang, the author of~\cite{Jiang2019Learning}, for helping us with the experiments of Soft3D. We also thank the strong support from Dr. Jinglei Shi, the author of~\cite{Shi2020Learning}, and his group. We acknowledge the \href{https://aimed.hku.hk}{Digital Health Laboratory} at the Department of Orthopaedics and Traumatology in the Faculty of Medicine, the University of Hong Kong for the supporting of model training and testing.

	
	%

	



	\ifCLASSOPTIONcaptionsoff
	\newpage
	\fi

	
	
	%
	\bibliographystyle{IEEEtran}
	\bibliography{IEEEabrv,transaction}

\end{document}